\documentclass[sigconf]{acmart}

\AtBeginDocument{%
  \providecommand\BibTeX{{%
    \normalfont B\kern-0.5em{\scshape i\kern-0.25em b}\kern-0.8em\TeX}}}

\usepackage{xurl}
\usepackage{multirow}
\usepackage{booktabs}
\usepackage{float}
\usepackage{hyperref}
\usepackage{subcaption}
\usepackage{amsmath}
\usepackage{mathtools}
\usepackage{amsfonts,amsmath,amsthm}
\newcommand{\svdots}{\raisebox{3pt}{$\scalebox{.75}{\vdots}$}} 
\newcommand{\sddots}{\raisebox{3pt}{$\scalebox{.75}{$\ddots$}$}} 
\newcommand{\indep}{\perp \!\!\! \perp}
\newtheorem{theorem}{Theorem}[section]

\begin{document}

\title{Dynamic Causal Structure Discovery and Causal Effect Estimation}

\author{Jianian Wang}
\affiliation{%
  \institution{North Carolina State University}
  \city{Raleigh}
  \state{North Carolina}
  \country{USA}
  \postcode{27606}
}
\author{Rui Song}
\affiliation{%
  \institution{North Carolina State University}
  \city{Raleigh}
  \state{North Carolina}
  \country{USA}
  \postcode{27606}
}

\begin{abstract}
  To represent the causal relationships between variables, a directed acyclic graph (DAG) is widely utilized in many areas, such as social sciences, epidemics, and genetics. Many causal structure learning approaches are developed to learn the hidden causal structure utilizing deep-learning approaches. However, these approaches have a hidden assumption that the causal relationship remains unchanged over time, which may not hold in real life. In this paper, we develop a new framework to model the dynamic causal graph where the causal relations are allowed to be time-varying. We incorporate the basis approximation method into the score-based causal discovery approach to capture the dynamic pattern of the causal graphs. Utilizing the autoregressive model structure, we could capture both contemporaneous and time-lagged causal relationships while allowing them to vary with time. We propose an algorithm that could provide both past-time estimates and future-time predictions on the causal graphs, and conduct simulations to demonstrate
the usefulness of the proposed method. We also apply the proposed method for the covid-data analysis, and provide causal estimates on how policy restriction’s
effect changes.
\end{abstract}




\maketitle

\section{Introduction}
To represent the causal relationships between variables, a directed acyclic graph (DAG) \citep{vanderweele} is widely utilized in many areas, such as social sciences, epidemics, and genetics \citep{pearl2009}. Based on the causal graph, we could calculate causal effects to illustrate the causal relations better. However, learning a faithful representation of the underlying causal graph is challenging, and there are many recent works that focus on the causal structure discovery problem.

Based on the fact that each graph could be scored using some score functions, such as BIC \citep{bic}, score-based approaches focus on finding the DAG which could yield the optimal score. For instance, 
\citet{GES} formulate the structure learning problem as a combinatorial optimizing problem and optimize the score by searching over the graph set with a greedy search method. Owing to the constraint that the graph must be acyclic, searching over the graph space is a combinatorial problem which is NP-hard \citep{nphard}, and the intractable problem poses challenges on optimization due to the computational difficulties in large-scale data. To overcome this challenge caused by acyclicity constraint, \citet{notears} convert the combinatorial optimization problem into a continuous optimization problem by setting the acyclicity constraint as a function of the adjacency matrix so that black-box solvers could be used to obtain the optimal graph efficiently. Motivated by this constraint setup, \citet{DAGGNN} apply the variational auto-encoder model to achieve better performance, and \citet{cai} provide a decomposition of the indirect effect, to categorize  interactions between mediators. 

Obtaining a causal graph using observations across time, the above literature often has a hidden assumption that the causal relation is fixed across time. However, the causal relations may not be static and may change with time. For instance, putting on a mask against wildfire smoke may be effective at the beginning, but will become less effective after long time of exposure \citep{wu}. Failing to capture the dynamic pattern of the causal graph, the conventional approaches may have large bias assuming the same causal structure among all observations.

However, modeling the causal relation in a dynamic way has many challenges. Firstly, to our best knowledge, the causal effect is not well-defined in the dynamic causal graphs, when the causal structure is subject to changes. Secondly, there are no off-the-shelf models on the time-varying causal graphs to represent the dynamic causal relations among the variables. Thirdly, allowing time-varying causal strength, i.e., weights that quantify the causal relations, may lead to the curse of dimensionality, since the complexity grows with the number of time stamps.

In this study, we develop a new framework to model the time-varying causal graph and propose definitions of the time-varying causal effect.
To capture the dynamic pattern of the causal graphs, we integrate the basis approximation method into the score-based causal discovery approach based on the dynamic linear structural equation model. By incorporating the autoregressive model structure, we are able to identify both contemporaneous and time-lagged causal relationships while allowing them to vary with time. We provide an algorithm that could estimate the causal graphs using historical information and also make accurate predictions for the future leveraging information spanning all time-periods. Our main contributions could be summarized as follows:

\begin{itemize}
    \item We propose the dynamic causal effect to represent treatment variable's effect on the outcome variable, providing a quantitative representation of time-varying causal relationships. The proposed dynamic causal effect is applicable to diverse scenarios where causal relations could be both time-varying and time-lagged.
    
    \item We propose a dynamic causal structure discovery approach which integrates the basis approximation method into the score-based method based on the proposed dynamic structural vector autoregression model. The proposed method has no implicit assumptions on the underlying causal strength functions and accommodates time-varying, time-lagged, and static causal relations.

    \item We propose an algorithm that could provide both past-time estimates and future-time predictions on the causal graphs, and conduct simulations to demonstrate the usefulness of the proposed method. We apply the proposed method to the real covid-data analysis and provide causal estimates on how policy restriction's effect changes. \footnote{The dataset and the code are publicly available at \url{https://github.com/jackie31425/Dynamic-Causal-Structure-Discovery-and-Causal-Effect-Estimation/}}
\end{itemize}

\section{Related work}
Our method is closely related to the causal structure learning literature. Three major types of methods are developed for causal structural learning to recover the hidden causal structure: constraint-based approach, structural equation models (SEM) based approach, and score-based approach \citep{survey}.

The first line of work, the constraint-based approach is based on the conditional independence constraint required by the Markov condition \citep{kiiveri}. A well-known example of the constraint-based approach is the PC algorithm \citep{PC}, which starts with a complete undirected graph and determines the skeleton and edge orientations by conducting conditional independence tests. Though the PC algorithm is shown to perform well under high-dimensional sparse graph \citep{PC_high_proof}, its output may vary if the order of the variables changes. The order dependence of the PC algorithm may result in non-stable results and \citet{PC_high} modify the PC algorithm and make it order-independent.

The second type of approach is the SEM-based approach, which imposes assumptions on data distribution while utilizing the structural equation models. Unlike the constraint-based approach, which often assumes Gaussian noise distribution, this line of method drops the Normal assumption and shows that the graph is identifiable under some conditions. ICA-LiNGAM \citep{shimizu}, for instance, assumes non-Gaussian noise to guarantee the model identification and independent component analysis (ICA) is utilized to obtain the linear model decomposition. Since the ICA method is an iterative search method, it may obtain an optimal local result instead of finding the global optima, and Direct-LiNGAM \citep{direct} is proposed to guarantee the convergence to the correct solution. 

The third major approach, the score-based approach, is based on the fact that each graph could be scored using some score functions, and it focuses on finding the DAG which could yield the optimal score. Greedy equivalence search (GES) \citep{GES}, for instance, performs a greedy search in the search space by comparing the scores after each addition and deletion. 
However, the previously mentioned scored-based approach optimizes the score over the graph set, and the optimization problem is NP-hard to solve due to its combinatorial nature. Instead, NOTEARS \citep{notears} converts the optimization problem to a continuous programming problem by formulating the acyclic constraint as a smooth equality constraint. By doing so, the DAG could be found by solving the programming problem with augmented Lagrangian. With the same spirit of forming the acyclic constraint as a function of adjacency matrix, DAG-GNN \citep{DAGGNN} applies a variational auto-encoder (VAE) to solve the optimization problem where encoders and decoders are parameterized using a graph neural network (GNN) structure to allow non-linear mappings. ANOCE \citep{cai} further extends DAG-GNN by incorporating an identification constraint based on background knowledge and enables the decomposition of the indirect effect. To take both contemporaneous and time-lagged causal relationships into consideration, \citet{dynotears} extend the NOTEARS approach by using the structural autogressive model, and \citet{graphnotears} further extend the model to handle the scenario where the replicates are no longer independent but have interactions.

Also, our method has a close connection to the dynamic causal effects, and there is some literature discussing similar concepts, including the time-varying treatment effect in longitudinal studies and dynamic Granger causality in time-series analysis. 

In longitudinal studies, literature aims to estimate the time-varying treatment effect based on the structural nested mean model (SNMM) \citep{robins1994}, where independent subjects are given treatments at each time point. For instance, \citet{robins1994} define the effect of the ”blip” of treatment variables, that is, holding a sequence of treatment variables at level 0, as the contrast for the respective potential outcomes conditionally on treatment and covariate histories. Instead of conditioning on all variable histories, \citet{boruvka} provide a generalization of the effect of the ”blip” of the treatment variable, so that it only conditions on a subset of variables and provides estimates using least square methods. The time-varying treatment effect is further extended by \citet{wu} to allow the causal effect to change with both time and locations, by proposing a spatially varying structural nested mean model. Without a structural equation model, this line of research cannot capture the inter-relations between mediators or the whole causal structure.

There is another line of work estimating time-varying Granger causality \citep{granger} in time-series analysis. Widely used in time-series analysis, Granger causality is defined if the predictability is improved by incorporating covariates and is commonly measured using the estimated coefficients. For instance, \citet{zhang} add time as a surrogate variable and apply kernel-based conditional independence test to estimate related coefficient. Since there may be time-lagged dependencies, \citet{huang} allow both causal strengths and noises variances to vary over time and use auto-regressive models to capture the time dependencies, while \citet{du} propose a hierarchical regression model to estimate the regression coefficient and the time-lag simultaneously. Based on the Granger causality, the literature above aims to estimate the time-varying coefficients but cannot directly provide estimates of the dynamic causal effects.
\section{Dynamic Causal Structure Discovery}
\subsection{Graph terminology}
\label{graph_terminology}
A graph $\mathcal{G}$ is defined by a pair: $\mathcal{G}=(\mathcal{V},\mathcal{E})$, where $\mathcal{V}=\{v_1,\cdots,v_p\}$ is a set of $p$ nodes and $\mathcal{E}$ is a set of edges, where $e_{ij}=(v_i,v_j)\in \mathcal{E}$ denotes an edge joining node $v_i$ and node $v_j$.
An adjacency matrix $\mathbf{A}$ is an $p \times p$ matrix, where $\mathbf{A}_{ij}$ represents the connection status between node $v_i$ and node $v_j$. 
 A dynamic graph is defined as a sequence of  graphs $\mathcal{G}^{seq}=\{\mathcal{G}_1,\cdots,\mathcal{G}_T\}$, where $ \mathcal{G}_i=(\mathcal{V}_i,\mathcal{E}_i), \text{for } i=1,\cdots,T$, $\mathcal{V}_i,\mathcal{E}_i$ are the set of nodes and edges for $i^{th}$ graph in the sequence respectively.

A path from $v_i$ to $v_j$ in $\mathcal{G}$ is a sequence of distinct vertices, \{$a_0, \cdots ,a_L\}\subset$ $\mathcal{V}$, such that $a_0 = v_i$, and $a_L = v_j$. A directed path from $v_i$ to $v_j$  is a path between $v_i$ and $v_j$  where all edges are directed toward $v_j$. A directed cycle is defined if there exist a
directed path from $v_i$ to $v_j$ and a directed path from $v_j$ to $v_i$ for any $v_i,v_j \in \mathcal{V}, i\ne j$. A directed acyclic graph is defined as a directed graph that does not contain directed cycles.

\subsection{Dynamic linear structural equation model}
\label{section_dlsem}
Specifying the relationship between variables through linear equations, the linear structural equation model \citep{lsem} is as follows:
\begin{align}
\mathbf{X}=\mathbf{X}\mathbf{B}+\mathbf{E}, 
\end{align}
where $\mathbf{X}\in \mathbb{R}^{m\times p}$ is a  data matrix for $m$ observation and $p$ variables, $\mathbf{B}\in \mathbb{R}^{p\times p}$ is an adjacency matrix of the DAG that characterizing the causal relationship of $\mathbf{X}$ and $\mathbf{E}\in \mathbb{R}^{m\times p}$ is the noise matrix.

Since the causal relations among the variables may not be static, we propose the following dynamic linear structural equation model (dynamic LSEM) to allow the causal structure to change with time:
\begin{align} 
   \mathbf{X}_{t}= \mathbf{X}_{t}\mathbf{B}_{t}+ \mathbf{E}_{t},
   \label{dynamic_lsem}
\end{align}
where $\mathbf{X}_t,\mathbf{B}_{t},\mathbf{E}_{t}$ denotes the $m\times p$ data matrix, $p\times p$ weighted adjacency matrix and $m\times p$ error matrix at $t^{th}$ timestamp. It is worth mentioning that we assume that there are no time-lagged dependencies in dynamic LSEM, and we will discuss a general model with time-lagged dependencies in Section \ref{section_lag}. 

\subsection{Identifiability}\label{identification}

In this section, we discuss the identifiability conditions.

\begin{theorem}
    Let $\mathcal{L}(\mathbf{X})$ denotes a function of $\mathbf{X}$ generated from equation \ref{dynamic_lsem}, 
 and $\mathcal{G}$ denotes the directed acyclic graph described in equation \ref{dynamic_lsem}. 
Under assumptions in Section \ref{assumption}, the graph $\mathcal{G}$ is identifiable from $\mathcal{L}(\mathbf{X})$ for the following two scenarios:
(1) The errors independently follow non-Gaussian distribution.
(2) The errors i.i.d. follows a Gaussian distribution.
\end{theorem}
These two conditions are commonly used in the literature \citep{nonpara_notears,shimizu} and in the following sections, we assume that one of these two conditions
on $\mathbf{E}_{t}$ holds.\footnote{The proof could be found in Appendix, Section \ref{section_proof_1}.}

\subsection{Basis approximation for varying coefficient modeling} \label{section_basis}


The dynamic linear structural equation model could be seen as a varying coefficient model \citep{hastie}, where the coefficient $\mathbf{B}_t$ is a function of time $t$. Nonparametrically modeling the coefficients may lead to the curse of dimensionality, since the complexity grows when the number of time stamps is large \citep{fan}. Instead, due to its fast convergence property, we utilize the basis approximation method \citep{huang_vary} to estimate the varying coefficient model, which conducts global smoothing on the coefficients. Then, each element of the coefficient matrix could be formulated as follows:
\setlength{\belowdisplayshortskip}{0pt}
\setlength{\abovedisplayshortskip}{0pt}
\begin{align}
    \begin{split}
   B_{ab,t}& \simeq\sum_{k=1}^{K}F_{k}(t)\gamma_{abk},\\ & a=1,\cdots,p, \quad b=1,\cdots,p , \quad t=1,\cdots, T,
   \end{split}
   \label{Bab}
\end{align}
where $B_{ab,t}$ denotes the $(a,b)^{th}$ element of the weighted adjacency matrix $B_t$, $t$ represents $t^{th}$ timestamp, 
$F_{k}(\cdot)$ is $k^{th}$ basis function, $K$ is the number of basis, $p$ is the number of covariates, $T$ is the number of timestamps and $\gamma_{abk}$ are the respective coefficients of $B_{ab,t}$ for $k^{th}$ basis. \footnote{Note that although the number of basis is finite in the current implementation, the approximation error may be ignorable since the number of basis is close to the maximum number of basis, which is equal to the length of the time span. Thus, we utilize equal signs in the following discussions.}

Then the dynamic LSEM could be written as 

\begin{align}
    \begin{split}
    \mathbf{X}_t&=(\mathbf{F}_t\otimes\mathbf{X}_t)\boldsymbol{\Gamma} +\mathbf{E}_t\\
    &=\mathbf{D}_t \boldsymbol{\Gamma} +\mathbf{E}_t,
    \end{split}
    \label{lsem_final}
\end{align}  

 $\text{where } t=1,\cdots, T,\mathbf{D}_t=\mathbf{F}_t\otimes\mathbf{X}_t\in \mathbb{R}^{m\times pK}, \otimes$ denotes the Kronecker product, $\mathbf{F}_t =[F_1(t),\cdots,F_K(t)]\in \mathbb{R}^{1\times K}$, and $\mathbf{\Gamma}\in \mathbb{R}^{pK\times p}$ contains the basis coefficient $\gamma_{abk}$.

Generally, we could express the data matrix  as
\begin{align}
    \mathbf{X}=\mathbf{D} \boldsymbol{\Gamma} +\mathbf{E}, \label{lsem}
\end{align}
where $\mathbf{X}=\begin{bmatrix}
    \mathbf{X}_{1} \\ \vdots \\\mathbf{X}_{T} 
\end{bmatrix}$, $\mathbf{D}=\begin{bmatrix}
    \mathbf{D}_{1} \\ \vdots \\\mathbf{D}_{T} 
\end{bmatrix}$ and $\mathbf{E}=\begin{bmatrix}
    \mathbf{E}_{1} \\ \vdots \\\mathbf{E}_{T} 
\end{bmatrix}$.

\subsection{Constrained causal structural learning} \label{algorithm}
A natural candidate to solve the basis approximation of the dynamic LSEM is the regression-based method. However, the regression-based method often imposes assumptions on the noise distribution, and this linear structure may fail to capture more complex data distribution. To overcome these limits, we utilize a variational auto-encoder (VAE)\citep{vae} with multi-layer perceptions (MLP) to generalize the LSEM structure for non-linear scenarios \footnote{The details of the VAE could be found in Appendix \ref{vae}}.

Specifically, we could reformulate Equation \eqref{lsem} as follows:
\vspace{-0.3cm} 
\begin{align*}
   \begin{bmatrix} \mathbf{X} & \mathbf{D} \end{bmatrix} &= \begin{bmatrix} \mathbf{X}_{m\times p} & \mathbf{D}_{m\times pK} \end{bmatrix} \begin{bmatrix} \mathbf{0}_{p\times p} & \mathbf{0}_{p\times pK} \\ \boldsymbol{\Gamma}_{pK\times p} & \mathbf{0}_{pK\times pK} \end{bmatrix}+\begin{bmatrix} \mathbf{E}_{m\times p} & \mathbf{E}'_{m\times pK} \end{bmatrix},\\
   &=
\begin{bmatrix} \mathbf{E}_{m\times p} & \mathbf{E}'_{m\times pK} \end{bmatrix} \begin{bmatrix} \mathbf{I}_{p\times p} & \mathbf{0}_{p\times pK} \\ -\boldsymbol{\Gamma}_{pK\times p} & \mathbf{I}_{pK\times pK} \end{bmatrix}^{-1},
\end{align*}

where $[\mathbf{E},\mathbf{E}']$ is a noise matrix which are then treated as latent variables in the variational autoencoder and MLP layers are utilized in both encoder and decoder structure.

To generate an estimate for the time-varying causal graph, we utilize evidence lower bound ($L_{ELBO}$) \citep{elbo} as the objective function, since the marginal log-likelihood is intractable due to the unknown posterior density. Furthermore, to guarantee that estimated graphs are valid causal graphs, we impose two constraints during the optimization process.

In order to ensure that the obtained graph at each time point is a DAG, we impose the acyclic constraint \citep{notears}: 
\vspace{-0.2cm} 
\setlength{\belowdisplayskip}{0pt}
\setlength{\abovedisplayskip}{0pt}
\begin{align}
    h_1(\boldsymbol{\Gamma})=\sum_{t=1}^T\{|tr[(\mathbf{I}_{p} + {\alpha} \mathbf{B}_t\odot \mathbf{B}_t)^{p}]-p|\} =0,
    \label{acyclic}
\end{align} 

where $tr(\cdot)$ represents the trace of the matrix, $\odot$ is the element-wise multiplication, $\mathbf{B}_t$ could be written as a function of $\boldsymbol{\Gamma}$ as in Equation \eqref{Bab}, $p$ is the number of variables, $T$ is the number of time-stamps and ${\alpha}$ is a hyperparameter.

As required by the assumptions in Section \ref{assumption}, the treatment variable has no parent nodes and the outcome variable has no child nodes, we propose the following treatment constraint,  
\begin{align}
h_2(\boldsymbol{\Gamma})=\sum_{i=1}^{pK}|\boldsymbol{\Gamma}_{i,0}|+\sum_{i=1}^{K}\sum_{j=1}^{p}|\boldsymbol{\Gamma}_{i*p,j}|=0,
\end{align}
where $\boldsymbol{\Gamma}_{i,j}$ is the $(i,j)^{th}$ element for matrix $\boldsymbol{\Gamma}$.

Thus, we could obtain the underlying causal graph by solving 
\begin{align*}
   \vspace{-0.5cm}  
\begin{cases}
    \min\limits_{\boldsymbol{\Gamma}} -L_{ELBO}
    &=-\frac{1}{p}\sum_{i=1}^p[
    -D_{KL}\{q(\mathbf{e}_i|\mathbf{x}_i)\Vert p(\mathbf{e}_i)\}\\
    &\qquad \qquad-\mathbb{E}_{q(\mathbf{e}_i|\mathbf{x}_i)}\{\text{log } p(\mathbf{x}_i|\mathbf{e}_i)
    \}],
\\
   \text{subject to }  & h_1(\boldsymbol{\Gamma})=0 \text{ and } h_2(\boldsymbol{\Gamma})=0.
    \end{cases}
\end{align*}

where $D_{KL}$ represents the KL divergence, $q(\mathbf{e}_i|\mathbf{x}_i)$ denotes the variational posterior of $\mathbf{e}_i$, $p(\mathbf{e}_i)$ denotes the prior distribution of $e_i$, $\text{log } p(\mathbf{x}_i|\mathbf{e}_i)$ denotes the log-likelihood function, $\mathbf{e}_i$ denotes the $i^{th}$ latent variable, and $\mathbf{x}_i$ denotes the $i^{th}$ data vector.

\section{Dynamic Causal Structure Discovery with Time-Lagged Dependency}
\label{section_lag}
\subsection{Dynamic structural vector autoregression}
In Section \ref{section_dlsem}, we propose the dynamic LSEM to allow the causal relations to vary with time, assuming that there are no temporal causal dependencies so that only same-time information would be causally related. However, this assumption may not hold in real-life, since there may exist time-lagged dependencies where the past observations could causally affect the current observations.

 To take the both time-lagged dependency and time-varying causal structure into account, we propose the following dynamic structural vector autoregressive (dynamic SVAR) model \citep{svar}:


\setlength{\belowdisplayskip}{0pt} \setlength{\belowdisplayshortskip}{0pt}
\setlength{\abovedisplayskip}{0pt} \setlength{\abovedisplayshortskip}{0pt}
\begin{align}
\mathbf{X}_t=\mathbf{X}_t\mathbf{B}_t+\mathbf{Z}_t\mathbf{W}_t+\mathbf{E}_t, \label{equation_SVAR}
\end{align}
where $\mathbf{X}_t\in \mathbb{R}^{m\times p}$ is the data matrix for $m$ observation and $p$ variables at $t^{th}$ time stamp, $\mathbf{B}_t\in \mathbb{R}^{p\times p}$ that characterizes the causal relationship of $X_t$, $\mathbf{Z}_t \in \mathbb{R}^{m\times pd}$ denotes the time-lagged data matrix for $\mathbf{X}_t$, $d$ denotes the order of the autoregressive model, $\mathbf{W}_t \in \mathbb{R}^{pd\times p}$ denotes the weight matrix that characterizing the time-lagged causal dependencies at $t^{th}$ time stamp and $\mathbf{E}_t\in \mathbb{R}^{m\times p}$ is the noise matrix at $t^{th}$ time stamp.
\subsection{Basis approximation and constrained causal structural learning}
Similar to Section \ref{section_basis}, we could utilize the basis approximation approach to solve the curse of the dimensionaty problem and obtain the estimates for the causal strength.\footnote{Detailed derivation could be found in Appendix Section \ref{4.2 derivation}.} \footnote{Similar to Section \ref{identification}, the graph is identifiable when the error under the conditions listed in Section \ref{identification}, since $\mathbf{B_t}$ is a DAG which satisfies the order condition for vector autoregressive models \citep{kilian}. }

Applying basis approximation, Equation \eqref{equation_SVAR} could be written as
\begin{align}
\begin{split}
    \mathbf{X}_t&=\mathbf{X}_t\mathbf{B}_t+\mathbf{Z}_t\mathbf{W}_t+\mathbf{E}_t, \\
    &=(\mathbf{F}_t\otimes\mathbf{X}_t)\boldsymbol{\Gamma}+(\mathbf{F}_t\otimes\mathbf{Z}_t)\mathbf{T}+\mathbf{E}_t, \\
    &=\mathbf{D}_t\boldsymbol{\Gamma}+\mathbf{G}_t\mathbf{T}+ \mathbf{E}_t,
    \label{equation_final}
\end{split} 
\end{align} 
$\text{where } t=d+1,\cdots, T, \otimes$ denotes the Kronecker product,
$\mathbf{F}_t =[F_1(t),\cdots,F_K(t)]\in \mathbb{R}^{1\times K}, \mathbf{D}_t=\mathbf{F}_t\otimes\mathbf{X}_t\in \mathbb{R}^{m\times pK},\mathbf{G}_t=\mathbf{F}_t\otimes\mathbf{Z}_t\in \mathbb{R}^{m\times pdK}$,
$\mathbf{\Gamma}\in \mathbb{R}^{pK\times p},\mathbf{T}\in \mathbb{R}^{pdK\times p}$
contains the basis coefficient respectively,

Stacking all the time points, we could have
\begin{align}
    \mathbf{X}=\mathbf{D}\boldsymbol{\Gamma}+\mathbf{G}\mathbf{T}+ \mathbf{E},
\end{align}
where $\mathbf{X}=\begin{bmatrix}
    \mathbf{X}_{d+1} \\ \vdots \\\mathbf{X}_{T} 
\end{bmatrix}$, $\mathbf{D}=\begin{bmatrix}
    \mathbf{D}_{d+1} \\ \vdots \\\mathbf{D}_{T} 
\end{bmatrix}$, $\mathbf{G}=\begin{bmatrix}
    \mathbf{G}_{d+1} \\ \vdots\\ \mathbf{G}_{T}\end{bmatrix} $ and $\mathbf{E} =\begin{bmatrix}
    \mathbf{E}_{d+1} \\ \vdots \\\mathbf{E}_{T} 
\end{bmatrix}$.\\

Then, Equation \eqref{equation_final} could be formulated as follows 
 \vspace{-0.1cm} 
\begin{align*}
   \begin{bmatrix} \mathbf{X} & \mathbf{D} & \mathbf{G}\end{bmatrix} &=  \begin{bmatrix} \mathbf{X} & \mathbf{D} & \mathbf{G}\end{bmatrix} \begin{bmatrix} \mathbf{0}_{p\times p} & \mathbf{0}_{p\times pK} & \mathbf{0}_{p\times pdK}\\ \boldsymbol{\Gamma} & \mathbf{0}_{pK\times pK}& \mathbf{0}_{pK\times pdK} \\ \mathbf{T} & \mathbf{0}_{pdK\times pK}& \mathbf{0}_{pdK\times pdK}\end{bmatrix}\\
   &\qquad +[\mathbf{E} \quad  \mathbf{E}'],\\
   &=
\begin{bmatrix} \mathbf{E}&\mathbf{E}' \end{bmatrix} \begin{bmatrix} \mathbf{I}_{p\times p} & \mathbf{0}_{p\times pK} & \mathbf{0}_{p\times pdK}\\ -\boldsymbol{\Gamma} & \mathbf{I}_{pK\times pK}& \mathbf{0}_{pK\times pdK} \\ -\mathbf{T} & \mathbf{0}_{pdK\times dK}& \mathbf{I}_{pdK\times pdK}\end{bmatrix}^{-1},
\end{align*}

where $[\mathbf{E} \quad \mathbf{E}']$ are noise matrices that are treated as latent variables in the variational autoencoder structure .

Since the future observations won't affect the past observations, there won't be edges pointing from the future data to the past data, which guarantees the acyclicity of $\mathbf{W}_t$. Thus it is sufficient to require $\mathbf{B}_t$ to be acyclic to guarantee the acyclicity of the entire causal graph. Thus, we could use Equation \eqref{acyclic} to impose the acyclicity constraint. 

Also, as required by the assumptions in Section \ref{assumption}, the treatment won't be affected by other variables while the outcome variable won't affect the other variables in the same time stamp, the treatment constraint is now:
   \vspace{-0.1cm}  
\begin{align}
    h_2^*(\boldsymbol{\Gamma}, \boldsymbol{T})=\sum_{i=1}^{pK}|\boldsymbol{\Gamma}_{i,0}|+\sum_{i=1}^{K}\sum_{j=1}^{p}|\boldsymbol{\Gamma}_{i*p,j}|+\sum_{i=1}^{pdK}|\boldsymbol{T}_{i,0}|=0,
\end{align}
where $\boldsymbol{\Gamma}_{i,j}, \boldsymbol{T}_{i,j}$ are the $(i,j)^{th}$ element for matrix $\boldsymbol{\Gamma},\boldsymbol{T}$ respectively.

Similar to Section \ref{algorithm}, we could obtain the estimated causal graph by solving
\vspace{-0.2cm} 
\begin{align*}
\begin{cases}
    \min\limits_{\boldsymbol{\Gamma},\boldsymbol{T}}& -L_{ELBO}\\
    &=-\frac{1}{p}\sum_{i=1}^p[
    -D_{KL}\{q(\mathbf{e}_i|\mathbf{x}_i)\Vert p(\mathbf{e}_i)\}\\
    &\qquad \qquad-\mathbb{E}_{q(\mathbf{e}_i|\mathbf{x}_i)}\{\text{log } p(\mathbf{x}_i|\mathbf{e}_i)
    \}],
\\
   \text{subject to }  & h_1(\boldsymbol{\Gamma})=0 \text{ and } h_2^*(\boldsymbol{\Gamma},\boldsymbol{T})=0.
    \end{cases}
\end{align*}
where $D_{KL}$ represents the KL divergence, $q(\mathbf{e}_i|\mathbf{x}_i)$ denotes the variational posterior of $\mathbf{e}_i$, $p(\mathbf{e}_i)$ denotes the prior distribution of $e_i$, $\text{log } p(\mathbf{x}_i|\mathbf{e}_i)$ denotes the log-likelihood function, $\mathbf{e}_i$ denotes the $i^{th}$ latent variable, and $\mathbf{x}_i$ denotes the $i^{th}$ data vector.
\section{Dynamic causal effect}
\subsection{Assumptions}
\label{assumption}
As commonly-used in the causal discovery literature \citep{PC,cai}, in this work, we have made the following assumptions.

Without loss of generality, we could decompose the data matrix $X_t$ into the treatment variable, $A_t$, mediator variable $M_t$ and the outcome variable $Y_t$, $X_t=[A_t,M_t^T,Y_t]$
Let overbar denotes the history of the respective variables, e.g. $\overline{A}_t=(A_1,\cdots, A_t)$, $Y_t{(\overline{a}_{t-1})}$ denote the potential outcome at time $t$ if the individual had the treatment history $\overline{a}_{t-1}$, and $H_t$ denotes the history up to time $t$, $H_t=(\overline{M}_{t},\overline{Y}_{t},\overline{A}_{t})$.

\noindent \textbf{Causal sufficiency} A set $\mathcal{V}$ of variables is causally sufficient for a population if and only if in the population every common cause of any two or more variables in $\mathcal{V}$ is in $\mathcal{V}$, or has the same value for all units in the population. \\
\noindent \textbf{Causal Markov condition} Each vertex is independent of its non-descendants in the graph conditional on its parents in the graph. In other words, we have 
\setlength{\belowdisplayskip}{0pt} \setlength{\belowdisplayshortskip}{0pt}
\setlength{\abovedisplayskip}{0pt} \setlength{\abovedisplayshortskip}{0pt}
\begin{equation}
   P(v_1, v_2, ... , v_n) =\prod_{i=1}^p P(v_i|pa(i)), \label{markov} 
\end{equation}
for vertices $\{v_1,...,v_p\}\in \mathcal{V}$, where $pa(i)$ are the parent nodes for $v_i$ and $P(\cdot)$ is a probability function .

\noindent \textbf{Consistency}: The observed data  are equal to the potential outcomes as follows:

$Y_{t} = Y_{t} (\Bar{A}_{t}),M_{t} = M_{t} (\Bar{A}_{t}),A_{t} = A_{t} (\Bar{A}_{t})$, for each $t\leq T$

\noindent \textbf{Sequential ignorability}: For each $t\leq T$, the potential outcome
$Y_{t} = Y_{t} (\Bar{A}_{t})$,  are independent of $A_t$ conditional on $H_t$ .

Based on the sequential ignorability assumption, the underlying treatment probabilities $p_t (1 | H_t ), t = 1,\cdots, T$, are some unknown constants and thus the treatment variable $A_t$ would have no parent node. Also, based on the consistency assumption, the outcome variable won't affect mediator variables at the same time-stamp.

\subsection{Dynamic causal effect} \label{section_causal_effect}
In this section, we propose the dynamic causal effect based on the dynamic structural vector autoregression model (dynamic SVAR), assuming the causal sufficiency, causal faithfulness, consistency, and sequential ignorability assumptions. \footnote{Note that the dynamic SVAR model would reduce to the dynamic LSEM model when the order $p=0$, so that the dynamic causal effect could be applied in both scenarios.}

Without loss of generality, we could decompose the data matrix $X_t$ into the treatment variable, $A_t$, mediator variable $M_t$ and the outcome variable $Y_t$, $X_t=[A_t,M_t^T,Y_t]$ and then Equation \eqref{equation_SVAR} could be written as:

\begin{equation}
\begin{split}
&[A_t,M_t^T,Y_t] \\
&= [A_t,M_t^T,Y_t] \mathbf{B}_{t}+ \mathbf{E}_{t}\\
&\qquad + [A_{t-1},M_{t-1}^T,Y_{t-1},\cdots,A_{t-d},M_{t-d}^T,Y_{t-d}] \mathbf{W}_{t}  \\
&= 
[A_t,M_t^T,Y_t]
\begin{bmatrix}
0 & \boldsymbol{\alpha}_{t} & \gamma_{t} \\ \boldsymbol{0}_{(p-2)\times 1} & \mathbf{C}_{t} & \boldsymbol{\beta}_{t} \\0  & \boldsymbol{0}_{1\times(p-2)} & 0 \end{bmatrix}+ \mathbf{E}_{t}\\
&\qquad +\sum_{i=1}^{d}
[A_{t-i},M_{t-i}^T,Y_{t-i}]
\begin{bmatrix}
0 & \boldsymbol{\alpha}_{t-i} & \gamma_{t-i} \\ \boldsymbol{0}_{(p-2)\times 1} & \mathbf{C}_{t-i} & \boldsymbol{\beta}_{t-i} \\0  & \mathbf{d}_{t-i}  & f_{t-i} \end{bmatrix},\\   
\end{split} \label{equation_amt}
\end{equation}

where $\gamma_{t},f_{t-1},\gamma_{t-i}$ are scalars, $ \boldsymbol{\alpha}_{t}^T,\boldsymbol{\beta}_{t},\boldsymbol{0}_{(p-2) \times 1}, \mathbf{d}_{t-i}^T$ are $(p-2)\times 1$ vectors, $\mathbf{C}_{t}$ is a $(p-2) \times (p-2)$ matrix, and $\mathbf{E}_{t}=[\epsilon_{A_{t}} , \epsilon_{M_{t}} , \epsilon_{Y_{t}}]$ is the error matrix. Based on the assumption, we could know that the treatment variable has no parent node and the outcome variable has no child node at the same time-stamp, which results in the zeros in the weight matrix.

Based on the literature \citep{boruvka, wu,robins1994,vanderweele}, we propose the dynamic causal effect as follows:

\begin{theorem}
Under assumptions in Section \ref{assumption} and Equation \eqref{equation_amt}, we could have the following dynamic causal effect:
\begin{align*}
&\mathbb{E}[Y_{t+1}{(\overline{a}_{t},a)}|H_t)
    -\mathbb{E}[Y_{t+1}{(\overline{a}_{t},0)}|H_t)\\
    &=\mathbb{E}[Y_{t+1}{(\overline{a}_{t},a)}|\overline{M}_{t}=\overline{m}_{t},\overline{Y}_{t}=\overline{y}_{t},\overline{A}_{t}=\overline{a}_{t}])
    -\\
    &\qquad \mathbb{E}[Y_{t+1}{(\overline{a}_{t},0)}|\overline{M}_{t}=\overline{m}_{t},\overline{Y}_{t}=\overline{y}_{t},\overline{A}_{t}=\overline{a}_{t}])\\
    &=(\gamma_{t+1}+\boldsymbol{\beta}_{t+1}^T(\mathbf{I}-\mathbf{C}_{t+1}^T)^{-1}\boldsymbol{\alpha}_{t+1}^T)a,
\end{align*}
\end{theorem}
where overbar denotes the history of the respective variables, e.g. $\overline{A}_t=(A_1,\cdots, A_t)$, $Y_t{(\overline{a}_{t-1})}$ denote the potential outcome at time $t$ if the individual had the treatment history $\overline{a}_{t-1}$, $a$ is a treatment value and $H_t$ denotes the history up to time $t$, $H_t=(\overline{M}_{t},\overline{Y}_{t},\overline{A}_{t}).$ The proof could be found in the Appendix Section \ref{section_proof}.
\section{Experimental results}

\subsection{Simulation studies}
\subsubsection{Basis selection}
To obtain an accurate basis approximation for the varying coefficient model, it is essential to select a basis system and determine the number of basis. 

In the following experiments, we utilize the B-spline as a basis system, since it has local support which is shown to improve the computational efficiency \citep{spline}. 

As commonly used in the literature, we utilize the order-2 B-spline with equally spaced knots, where the number of knots is selected using cross-validation \citep{huang_vary}. \footnote{Details of the selected B-spline could be found in the Appendix Section \ref{spline_detail}. }

\subsubsection{Dynamic linear structural equation model (dynamic LSEM) setup} The data are generated based on Equation \eqref{dynamic_lsem} with Gaussian errors, where each graph has 5 variables ($p=5$) \footnote{The graph size is set to match the graph size of the real-data. We have provided evaluations on varying graph size in Appendix Section \ref{addtional}.} and 30 observations ($m=30$). We consider 10 time stamps ($T=10$) and generate 30 realizations. 

We consider two scenarios of graph: 
\begin{itemize}
    \item (S1) The true underlying causal graph has only one edge ($A\to Y$), and the strength of the causal relation change with time; \\
    \itemsep -1em
    \item  (S2) The true underlying causal graph is generated from the Erdos-Renyi model with an expected degree as 4 and the strength of the causal relations are randomly assigned to be time-varying or static.
\end{itemize}

We consider two different functions for the causal relations: 
\begin{itemize}
    \item (F1) Cosine function: $f_1(t)=cos(\frac{t}{4\pi})*0.8;$\\
    \itemsep -1em
    \item (F2) Quadratic function: $f_2(t)= \frac{-10+(5-t)^2}{20}.$
\end{itemize}

\begin{figure}[h]
\begin{subfigure}{0.75\columnwidth}
  \centering
  \includegraphics[width=\columnwidth]{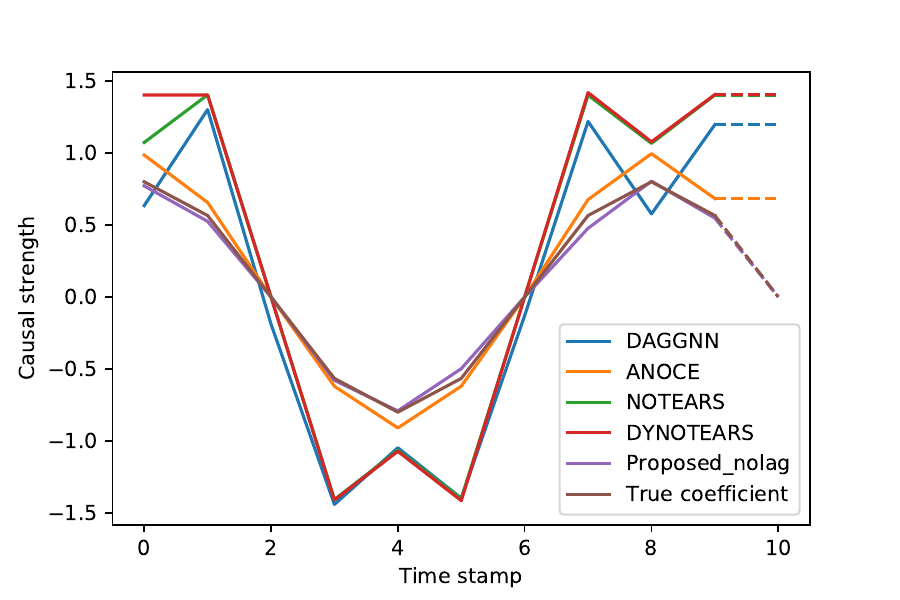}
  \caption{Cosine function F1}
\end{subfigure}%
\hfill
\begin{subfigure}{0.75\columnwidth}
  \centering
  \includegraphics[width=\columnwidth]{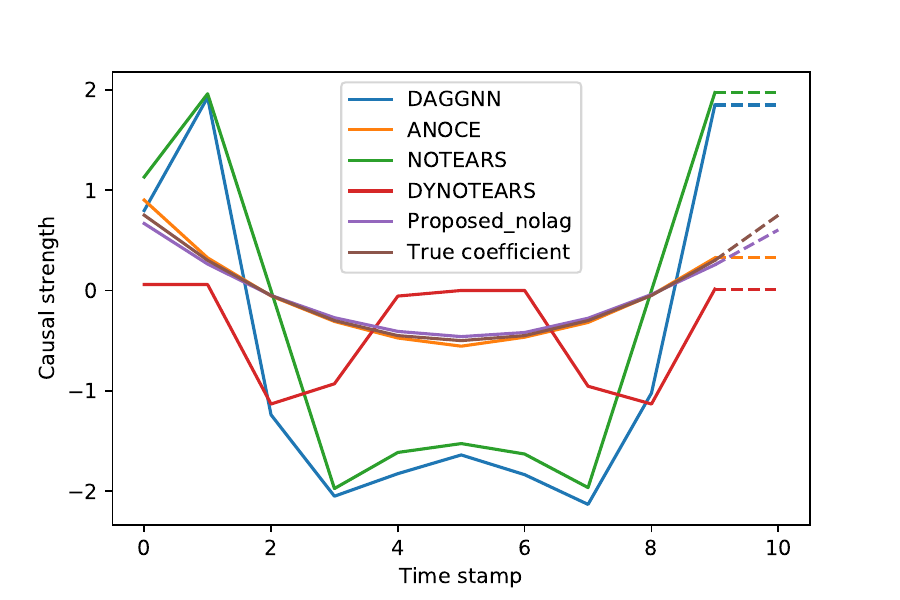}
  \caption{Quadratic function F2}
\end{subfigure}
\caption{Estimated causal strength 
for Scenario 1 in the dynamic LSEM setup. The solid line denotes the estimated coefficient in the ten time stamps and the dashed line denotes the one-step ahead prediction.}
\label{figure}
\end{figure}

\begin{table}[h]
{\centering
\caption{The empirical comparison of estimated contemporaneous weight matrix $\mathbf{B}_t$ in the dynamic SVAR model setup. The number in parenthesis denotes the standard deviation. }

\begin{center}
\begin{tabular}{c@{}cc@{}cc@{}c@{}}
\toprule
Methods & Metric & F1          & F2                     &  &  \\
\midrule
\multirow{4}{*}{DAGGNN}    & FDR    & 0.79(0.02)   & 0.80(0.01)       &  &  \\
& TPR    & 0.83(0.02)    & 0.76(0.02)    &  &  \\
 & SHD    & 3.05(0.13)    & 3.09(0.10)       &  &  \\
  & MSE    & 0.24(0.02)    & 0.24(0.02)       &  &  \\
\midrule
\multirow{4}{*}{ANOCE}    & FDR    & 0.77(0.01)   & 0.77(0.01)       &  &  \\
& TPR    & 0.82(0.02)    & 0.83(0.02)    &  &  \\
 & SHD    & 2.53(0.10)    & 2.61(0.10)       &  &  \\
& MSE    & 0.07(0.01),        & 0.10 (0.01)            &  &  \\
\midrule
\multirow{4}{*}{NOTEARS}    & FDR    & 0.36(0.03)   & 0.39(0.03)          &  \\
& TPR    & 0.61(0.02)    & 0.7(0.02)     &    \\
 & SHD    & 0.82(0.06)    & 0.80(0.06)        &    \\
& MSE    & 0.10(0.01)               & 0.13(0.01)        &    \\
\midrule
\multirow{4}{*}{DYNOTEARS} & FDR    & \textbf{0.05 (0.01)} & 0.31(0.01)     &    \\
& TPR    & \textbf{0.98(0.01)}    & 0.70(0.01)    &    \\
& SHD    & \textbf{0.05(0.01)}  & 0.32(0.01)       &    \\
& MSE    & 0.15(0.00)        & 0.40(0.01)              &   \\
\midrule
\multirow{4}{*}{Proposed}    & FDR    & 0.13 (0.02)   & \textbf{0.05 (0.01)}        &    \\
& TPR    & \textbf{0.98(0.00)}    & \textbf{0.98(0.01)}     &  & \\
 & SHD    & 0.32(0.05)   & \textbf{0.05(0.01)}       &    \\
& MSE    & \textbf{0.02(0.01)}        & \textbf{0.02 (0.02)}        &    \\

\bottomrule
\end{tabular}
\end{center}
\label{table_lag}
}

\end{table}
\subsubsection{Dynamic structural vector autoregression (dynamic SVAR) model setup} The data are generated based on Equation \eqref{equation_SVAR} with Gaussian errors, where each graph has 5 variables ($p=5$) and 30 observations ($m=30$). We consider 10 time stamps ($T=10$) and generate 30 realizations. The order of the autoregressive model is $d=1$ and the time-lagged weight matrix $\mathbf{W}_t=\left[\begin{smallmatrix}
    0&0&0&0&0\\0&1&0&0&0\\0&0&1&0&0\\0&0&0&1&0\\ 0&0&0&0&1
\end{smallmatrix}\right], \forall t.$
The contemporaneous weight matrix $\mathbf{B}_t$ has been designed with only one edge ($A\to Y$), whose weight changes with time as the following two different functions: 
\begin{itemize}
    \item (F1) Cosine function: $f_1(t)=cos(\frac{t}{4\pi})*0.8;$\\
    \itemsep -1em
    \item (F2) Quadratic function: $f_2(t)=\frac{-15+(5-t)^2}{25}.$
\end{itemize}

\subsubsection{Evaluation metric and benchmark methods}
To evaluate the performance of the proposed method, we utilize the proposed method to identify the hidden causal graph and compare it with the commonly used causal discovery approaches: DAGGNN \citep{DAGGNN}, NOTEARS \citep{notears}, DYNOTEARS \cite{dynotears} and ANOCE \citep{cai}. Since the benchmark approaches are developed for the static causal graph, for each timestamp, we generate estimates for the causal graph using the current timestamp's data and use the most-recent estimate as the one-step ahead prediction. We use all-time data to generate historical estimates and the future predictions using the proposed method \footnote{For each model setup, we utilize the respective proposed dynamic structure discovery method and label it as the proposed method.}. To better assess the the proposed models' ability to capture the dynamic pattern, we also utilize the CD-NOD \citep{zhang} as an benchmark, which could conduct causal discovery for non-stationary process.\footnote{The CD-NOD method is developed in the setup where there are no replicates and have long time-spans. Thus, we cannot apply it in the forementioned scenarios and tested it in the scenario where $m=1, T=100$. The result could be found in Appendix Section \ref{addtional}.}

We evaluate the graph estimates using the following three metrics: false discovery rate (FDR), true positive rate (TPR), and structural Hamming distance (SHD). To evaluate the time-varying strength of causal relations, we use mean squared error (MSE) as an evaluation metric. It is the higher the better for the TPR, while for the other three metrics, a lower value indicates a better result. As commonly used in the causal discovery literature \citep{cai,notears}, we remove the edges if the edge weight is lower than 0.2 to reduce noise. 
\subsubsection{Implementation details}
In this section, we present the implementation details for the proposed method and the benchmark methods as follows: 
\begin{itemize}
    \item The computation is done by the processer Intel(R) Core(TM) i7-8550U CPU. The dataset and the code are publicly available at the repository \url{https://github.com/jackie31425/Dynamic-Causal-Structure-Discovery-and-Causal-Effect-Estimation/settings}.
    \item Proposed method: The proposed method is implemented based on PyTorch \citep{pytorch} using Adam \citep{adam} optimizer to optimize the loss function. The batch size is set to be 10, training epoch is 200, and the number of hidden nodes is the square of the number of variables. The maximum iteration number for searching parameters is 100, the initial learning rate is 0.003 which decays  with a decay rate of 1.0.
    \item ANOCE \citep{cai}: ANOCE provides a decomposition of the indirect effect, to categorize interactions between mediators. We implement the ANOCE with the default hyper-parameters to estimate the causal graph at each time-stamp using each-times' data. Their code is available at the repository https://github.com/hengruicai/ANOCE-CVAE.
    \item NOTEARS \citep{notears} converts the combinatorial optimization
problem into a continuous optimization problem by
setting the acyclicity constraint as a function of the
adjacency matrix. We implement the NOTEARS with the default hyper-parameters to estimate the causal graph at each time-stamp using each-times' data. Their code is available at the repository https://github.com/xunzheng/notears.
\item DYNOTEARS \citep{dynotears} extends the NOTEARS model to handle the scenario where the replicates are no longer independent but have interactions. We implement the DYNOTEARS with the default hyper-parameters to estimate the causal graph at each time-stamp using lag-1 data. Their code is available at the repository https://github.com/mckinsey/causalnex/blob/ develop/causalnex/structure/dynotears.py.
\item CD-NOD \citep{zhang} proposes the constriant-based procedure to detect the changing causal structure. We implement the CD-NOD in Matlab \citep{matlab} with the default hyper-parameters and set the number of iterations to be 1000. Their code is available at the repository https://github.com/Biwei-Huang/Causal-Discovery-from-Nonstationary-Heterogeneous-Data.
    
\end{itemize}

\subsubsection{Results}
\begin{figure}[h]

\includegraphics[scale=0.25]{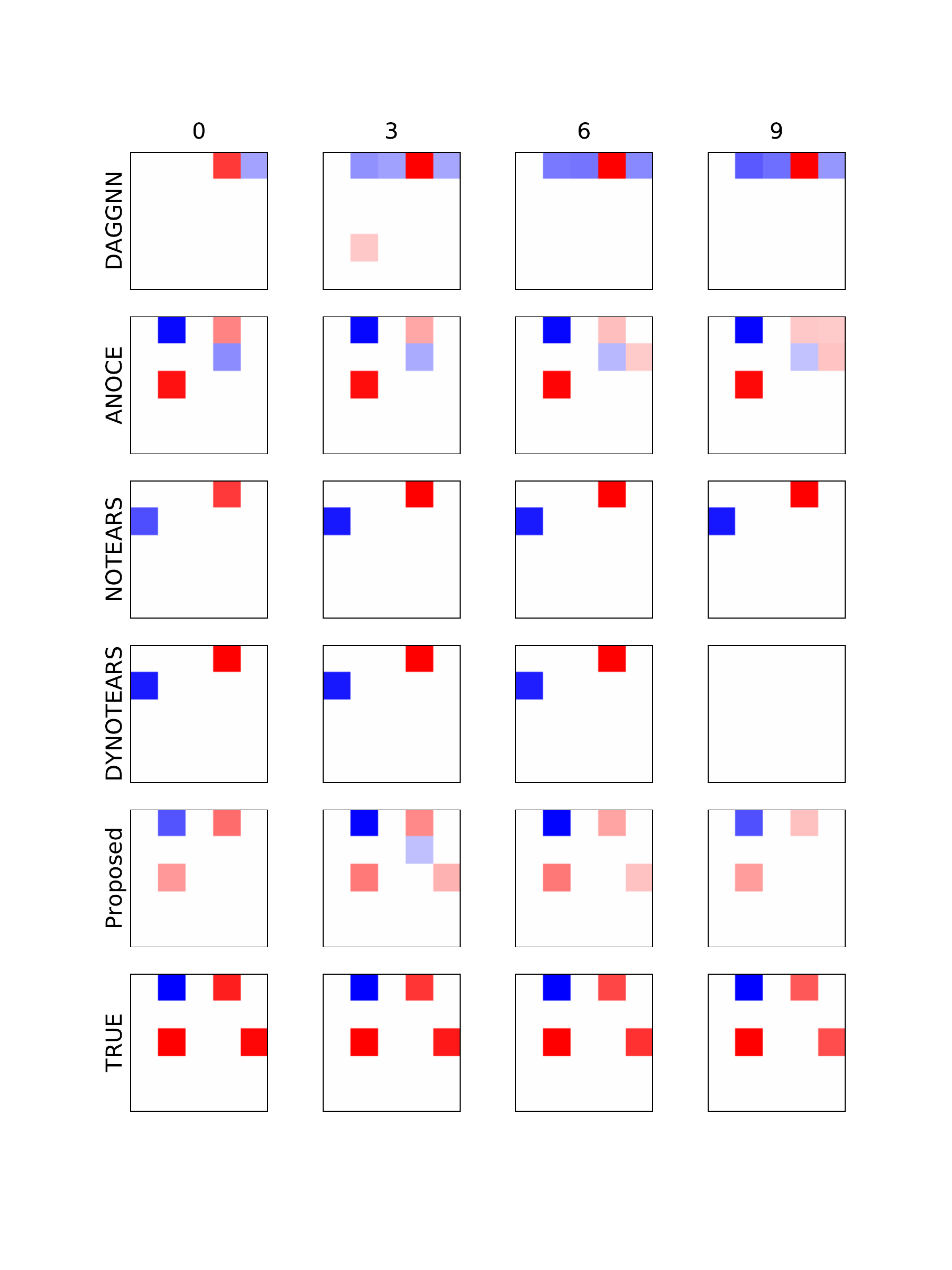}
\vspace*{-1.5cm}
    \caption{Estimated causal graphs at multiple time-stamps for dynamic LSEM setup, Scenario 2. The horizontal axis denotes the number of timestamps.}
\label{S2_nolag_graph}
\end{figure}

 The results for the proposed method and the benchmark methods are shown in Table \ref{table_lag},\ref{table}, \ref{table_m=1},\ref{table_lag_W_t},\ref{table_vary} and Figure \ref{figure},\ref{S2_nolag_graph},\ref{figure_lag},\ref{S1_nolag_graph},\ref{lag_graph} \footnote{The Figure \ref{figure_lag},\ref{S1_nolag_graph},\ref{lag_graph} and Table \ref{table_m=1}, \ref{table_lag_W_t},\ref{table_vary} could be found in Appendix Section \ref{addtional}.}. The findings are summarized as follows.

As shown in Figure \ref{figure}, \ref{figure_lag}, the proposed method could generate more accurate estimates for the causal strength in both cosine and quadratic function setups. Moreover, the proposed method is able to obtain an accurate prediction for the next time stamp, while the other benchmark methods fail to do so.

Also, as shown in Table \ref{table_lag},\ref{table},\ref{table_lag_W_t}  we could conclude that the proposed method could also capture the hidden graph structure since the proposed method obtains better results in terms of FDR, SHD, and MSE in most of the scenarios while having comparable TPR with benchmark methods. Furthermore, visual representations of the estimated graphs (Figure \ref{S2_nolag_graph},\ref{S1_nolag_graph},\ref{lag_graph}), also illustrate that the proposed method excels in estimating the causal graph structure. Moreover, as shown in Table \ref{table_vary}, the proposed method still have superior performance when the graph size increases.

\begin{table}[h]
\caption{The empirical comparison of the estimated causal graph for the synthetic data in the dynamic LSEM setup. The number in parenthesis denotes the standard deviation. }
\centering
\begin{tabular}{c@{}cc@{}cc@{}cc@{}}
\toprule
Methods & Metric & S1F1          & S1F2          & S2            &  &  \\
\midrule
\multirow{4}{*}{DAGGNN}    & FDR    & 0.83(0.01)   & 0.80(0.01)    & 0.84(0.03)    &  &  \\
& TPR    & 0.82(0.04)    & 0.90(0.03) & 0.48(0.03)    &  &  \\
 & SHD    & 3.36(0.16)    & 3.38(0.15)    & 5.10(0.17)    &  &  \\
& MSE    & 0.65(0.04),        & 2.69(0.26)        & 0.14(0.02)        &  &  \\
\midrule
\multirow{4}{*}{ANOCE}    & FDR    & 0.54(0.03)   & 0.54(0.03)    & 0.36(0.02)    &  &  \\
& TPR    & \textbf{1.00(0.00)}    & \textbf{1.00(0.00)} & 0.84(0.03)    &  &  \\
 & SHD    & 1.83(0.13)    & 1.81(0.13)    & 2.83(0.17)    &  &  \\
& MSE    & 0.06(0.02),        & \textbf{0.02 (0.01)}        & 0.15(0.01)        &  &  \\
\midrule
\multirow{4}{*}{NOTEARS}    & FDR    & \textbf{0.00(0.00)}   & \textbf{0.00(0.00)}    & 0.50(0.02)   &  &  \\
& TPR    & \textbf{1.00(0.00)}    & \textbf{1.00(0.00)} & 0.25(0.03)    &  &  \\
 & SHD    & \textbf{0.00(0.00)}    & \textbf{0.00(0.00)}    & 3.02(0.17)    &  &  \\
& MSE    & 0.37(0.00),        & 1.51 (0.01)        & 0.52(0.01)        &  &  \\
\midrule
\multirow{4}{*}{DYNOTEARS}    & FDR    & \textbf{0.00(0.00)}   & 0.50(0.04)    & 0.36(0.02)    &  &  \\
& TPR    & \textbf{1.00(0.00)}    & 0.51(0.04) & 0.25(0.00)    &  &  \\
 & SHD    & \textbf{0.00(0.00)}    & 0.70(0.06)   & 3.03(0.01)    &  &  \\
& MSE    & 0.66(0.00),        & 0.54(0.01)        & 0.50(0.01)        &  &  \\
\midrule
\multirow{4}{*}{Proposed} & FDR    & \textbf{0.00 (0.00)} & \textbf{0.00   (0.00)} & \textbf{0.12(0.01)}    &  &  \\
& TPR    & \textbf{1.00(0.00)}    & 0.94(0.03)   & \textbf{0.85   (0.02)} &  &  \\
& SHD    & \textbf{0.01(0.01)}  & 0.06(0.03)   & \textbf{1.12(0.11)}    &  &  \\
& MSE    & \textbf{0.00(0.00)}        & \textbf{0.01 (0.00)}        & \textbf{0.08(0.01)}         &  & \\
\bottomrule
\end{tabular}
\label{table}
\end{table}


\subsection{Real data analysis}
Starting in 2019, the coronavirus disease (covid-19) has spread globally and caused many human lives to be lost. There are many literature studying the factors that affect the covid-19 transmission, such as policy restriction \citep{covid-restriction}and people's awareness \citep{covid-search}. However, the effects of these factors may be time-varying. For instance, the contact restriction policy may be implemented for several months, but its effect may decrease since people may not obey the rules in the later stage. In this section, we aim to learn a dynamic causal graph based on the covid-19 data to study the effects of policy intervention on the covid-19 cases.


We collect weekly aggregated data on 27 districts in Germany from February 15 to July 8, 2020 \citep{steiger}. Summarized by \citet{steiger}, the possible factors for covid-19 could be categorized into five categories: mobility, awareness, weather, intervention, and socio-demographic factors. Since weather and socio-demographic factors will not be influenced by the policy intervention, we don't include these factors and select contact restriction policy as the treatment variable, average mobility and searches for corona as the mediator variables \footnote{The treatment variable is a binary variable, representing whether the contact restriction policy is implementing. Mobility measures how the community are moved based on Google community mobility reports \citep{google_mobility}. Searches represents the relative interest in the term "corona" in Google Search.}, and reported new cases as the outcome variable. The dataset has 4 variables ($p=4$), 27 observations ($m=27$) and 20 time stamps ($T=20$).  As shown in Figure \ref{real_data}, after the contact restriction policy beginning to implement, people tend to travel less since the mobility variable has a sudden drop. Also, the average reported new cases shows a decreasing trend after the policy begins to implement.

We apply the two proposed methods to the real data and compare their ELBO loss. Since it is shown in the literature that these factors may have a lagged effect of 5 days \cite{steiger}, the proposed method based on autoregressive model is applied with order $d=1$. This proposed model with time-lagged dependency has a smaller ELBO loss (0.02) \footnote{This proposed model with time-lagged dependency has the log-likelihood loss of 0.002 and MSE of $8.06*10^{-6}$.} compared to that of the model without time-lagged dependency (0.06), which also matches with the literature. Thus, we present the results of the proposed method that is based on autoregressive model, since it has better performance and more intuitive.

\begin{figure}[h]
\begin{subfigure}{\columnwidth}
  \centering
  \includegraphics[width=0.85\columnwidth]{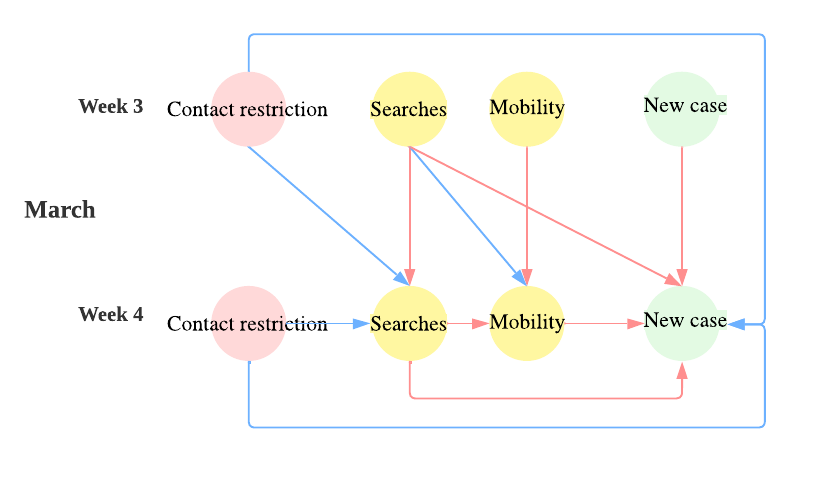}
  \caption{Estimated causal graph at March}
  \label{march}
\end{subfigure}%
\hfill
\begin{subfigure}{\columnwidth}
  \centering
  \includegraphics[width=0.85\columnwidth]{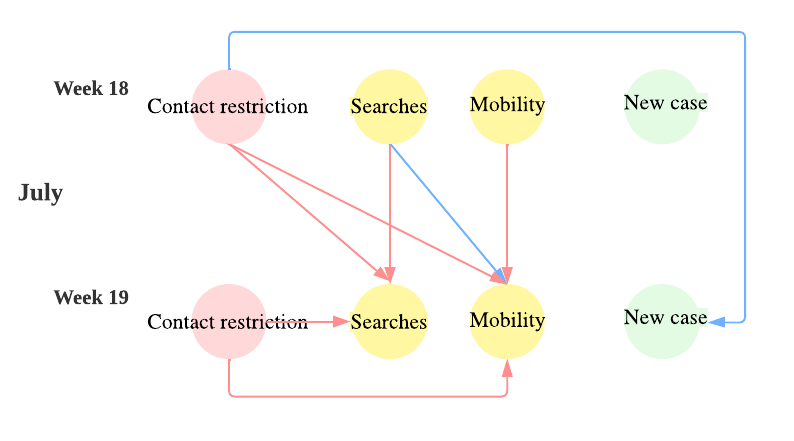}
  \vspace*{-0.5cm}
  \caption{Estimated causal graph at July}
  \label{july}
\end{subfigure}
\caption{Estimated causal graph at March and July. Each node represents a variable and the arrows represent the discovered causal relations. Red color represents positive causal relations and blue color represents negative causal relations. }
\label{graph_lag}
\end{figure}

Figure \ref{graph_lag} presents the estimated causal graph at March and July. We could see that in both graphs, the contact restriction policy could reduce new COVID cases for the next week and the mediator variables, i.e., average mobility and searches for corona, could positively affect themselves in the next week. We could also notice that, in March, implementing the contact restriction policy could indirectly reduce mobility as shown in Figure \ref{march}. However, in July, this policy doesn't have negative effects on mobility and may even increase it, as shown in Figure \ref{july}. It suggests that the policy may not be properly implemented, as people starting to going out more.

\begin{figure}[h]
    \centering
    \vspace*{-0.15in}
    \includegraphics[width=0.7\linewidth]{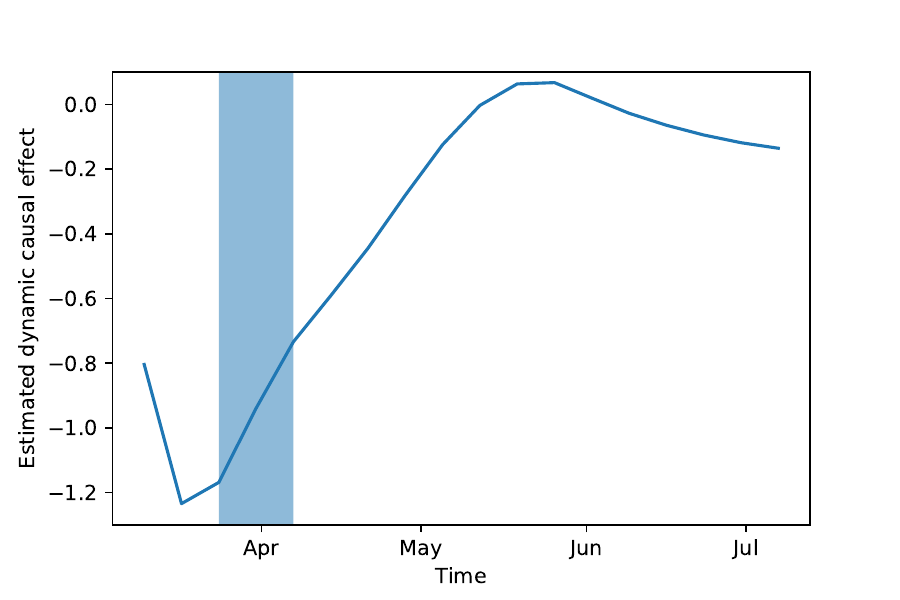}
    \caption{Estimated dynamic causal effect of the contact restriction policy on new cases. The shaded region represents the time period when the policy starts to implement.}
    \label{TE}
\end{figure}

Figure \ref{TE} presents the estimated dynamic causal effect of the contact restriction policy on the reported new cases, illustrating how the policy influences the number of new cases. The negative estimated effect suggests that implementing this policy can help reduce the spread of COVID-19. However, this effect diminishes over time, since the estimated effect becomes smaller in magnitude, possibly due to a decline in the effectiveness of policy implementation. Additionally, Figure \ref{TE} indicates that it might be more effective for the government to implement the contact restriction policy earlier, as the estimated dynamic causal effect is stronger in March.

\section{Conclusion and Discussion}
In this paper, we develop a new framework to model the dynamic causal graph where causal relations are allowed to be both time-varying and time-lagged. We propose a score-based dynamic causal structure discovery approach and propose an algorithm that could provide estimates on both dynamic casual graphs and dynamic causal effects. To conclude, we briefly discuss some limitations and possible future directions.

Firstly, we apply the basis approximation method to estimate the time-varying casual strength and may get less accurate results when the true relationship is not continuous. This issue may be resolved using more basis, but it may increase computation complexity. 

Secondly, the basis approximation method may lead to estimation error since the number of basis is finite in the implementation. It may be better to incorporate the basis estimation bias into the loss function to generalize the method to datasets with long time spans.

Thirdly, background information is utilized to determine the order of the autoregressive model, which may not available in some scenarios. A possible approach is to utilize the cross-validation method to determine the appropriate order, but it may also increase the computational time.

\clearpage
\bibliographystyle{ACM-Reference-Format}
\bibliography{ref}

\clearpage
\appendix

\section{Detailed derivation of basis approximation in Section 4.2}
\label{4.2 derivation}
We could represent each element of the $\mathbf{B}_t,\mathbf{W}_t$ as follows:
\begin{align}
\begin{split}
   B_{mn,t}&\simeq\sum_{k=1}^{K}F_{k}(t)\gamma_{mnk}, \\&\quad m=1,...,p, n=1,...,p, t=(d+1),..., T\\
   W_{mn,t}&\simeq\sum_{k=1}^{K}F_{k}(t)\tau_{mnk}, \\&\quad m=1,...,pd,  n=1,...,p, t=(d+1),..., T
   \end{split}
\end{align}
where $B_{mn,t},W_{mn,t}$ denotes the $(m,n)^{th}$ element of the causal strength matrix $\mathbf{B}_t,\mathbf{W}_t$, $t$ represents $t^{th}$ timestamp, $F_{k}(\cdot)$ is $k^{th}$ basis function, $K$ is the number of basis, $p$ is the number of covariates, $d$ denotes the order of the autoregressive model, $T$ is the number of timestamps and $\gamma_{mnk}, \tau_{mnk}$ are the respective coefficients of $B_{mn,t},W_{mn,t}$ for $k^{th}$ basis.

Then we could write $\mathbf{X}_t\mathbf{B}_t$ as follows
\begin{equation*}
\addtolength{\arraycolsep}{-5pt}
\begin{split}
    \mathbf{X}_t\mathbf{B}_t&=
\left[\begin{smallmatrix}
    X_{t11} & \cdots & X_{t1p}\\ \svdots & \sddots & \svdots \\X_{tm1} & \cdots & X_{tmp}
\end{smallmatrix}\right]
 \left[\begin{smallmatrix}
    \sum_{k=1}^{K}F_{k}(t)\gamma_{11k} & \cdots & \sum_{k=1}^{K}F_{k}(t)\gamma_{1pk}\\ \svdots & \sddots & \svdots  \\\sum_{k=1}^{K}F_{k}(t)\gamma_{p1k} & \cdots & \sum_{k=1}^{K}F_{k}(t)\gamma_{ppk}
\end{smallmatrix}\right]\\ &\qquad+ \mathbf{E}_t\\
&= \left[\begin{smallmatrix}
  F_{1}(t)X_{t11} & \cdots &  F_{1}(t)X_{t1p} & \cdots & F_{K}(t)X_{t11} & \cdots &  F_{K}(t)X_{t1p} \\
\svdots & \sddots & \svdots & \sddots & \svdots & \sddots & \svdots \\
  F_{1}(t)X_{tm1} & \cdots &  F_{1}(t)X_{tmp} & \cdots & F_{K}(t)X_{tm1} & \cdots &  F_{K}(t)X_{tmp}
\end{smallmatrix}\right]\\
&\qquad\left[\begin{smallmatrix}
\gamma_{111} & \cdots & \gamma_{1p1} \\ \svdots & \sddots & \svdots \\ \gamma_{p11} & \cdots & \gamma_{pp1} \\ \svdots & \sddots & \svdots \\ \gamma_{11K} & \cdots & \gamma_{1pK} \\ \svdots & \sddots & \svdots \\ \gamma_{p1K} & \cdots & \gamma_{ppK}\\
\end{smallmatrix}\right]+ \mathbf{E}_t,\\
&=\mathbf{D}_t\boldsymbol{\Gamma}+\mathbf{E}_t , \quad t=(d+1),\ldots, T,
\end{split}
\end{equation*}
where
\\ $\mathbf{D}_t=\left[\begin{smallmatrix}
  F_{1}(t)X_{t11} & \cdots &  F_{1}(t)X_{t1p} & \cdots & F_{K}(t)X_{t11} & \cdots &  F_{K}(t)X_{t1p} \\
\svdots & \sddots & \svdots & \sddots & \svdots & \sddots & \svdots \\
  F_{1}(t)X_{tm1} & \cdots &  F_{1}(t)X_{tmp} & \cdots & F_{K}(t)X_{tm1} & \cdots &  F_{K}(t)X_{tmp}
\end{smallmatrix}\right]$,\\
and $\boldsymbol{\Gamma}=\qquad\left[\begin{smallmatrix}
\gamma_{111} & \cdots & \gamma_{1p1} \\ \svdots & \sddots & \svdots \\ \gamma_{p11} & \cdots & \gamma_{pp1} \\ \svdots & \sddots & \svdots \\ \gamma_{11K} & \cdots & \gamma_{1pK} \\ \svdots & \sddots & \svdots \\ \gamma_{p1K} & \cdots & \gamma_{ppK}\\
\end{smallmatrix}\right]$.

Similarly, we could have $\mathbf{Z}_t\mathbf{W}_t$ as follows
\begin{equation*}
\addtolength{\arraycolsep}{-5pt}
\begin{split}
   &\mathbf{Z}_t\mathbf{W}_t\\
   &=
\left[\begin{smallmatrix}
    Z_{t11} & \cdots & Z_{t1(pd)}\\ \svdots & \sddots & \svdots \\Z_{tm1} & \cdots & Z_{tm(pd)}
\end{smallmatrix}\right]
 \left[\begin{smallmatrix}
    \sum_{k=1}^{K}F_{k}(t)\tau_{11k} & \cdots & \sum_{k=1}^{K}F_{k}(t)\tau_{1pk}\\ \svdots & \sddots & \svdots  \\\sum_{k=1}^{K}F_{k}(t)\tau_{(pd)1k} & \cdots & \sum_{k=1}^{K}F_{k}(t)\tau_{(pd)pk}
\end{smallmatrix}\right]\\ &\qquad+ E_t\\
&= \left[\begin{smallmatrix}
  F_{1}(t)Z_{t11} & \cdots &  F_{1}(t)Z_{t1(pd)} & \cdots & F_{K}(t)Z_{t11} & \cdots &  F_{K}(t)Z_{t1(pd)} \\
\svdots & \sddots & \svdots & \sddots & \svdots & \sddots & \svdots \\
  F_{1}(t)Z_{tm1} & \cdots &  F_{1}(t)Z_{tm(pd)} & \cdots & F_{K}(t)Z_{tm1} & \cdots &  F_{K}(t)Z_{tm(pd)}
\end{smallmatrix}\right]\\
&\qquad\left[\begin{smallmatrix}
\tau_{111} & \cdots & \tau_{1p1} \\ \svdots & \sddots & \svdots \\ \tau_{(pd)11} & \cdots & \gamma_{(pd)p1} \\ \svdots & \sddots & \svdots \\ \tau_{11K} & \cdots & \tau_{1pK} \\ \svdots & \sddots & \svdots \\ \tau_{(pd)1K} & \cdots & \tau_{(pd)pK}\\
\end{smallmatrix}\right]+ \mathbf{E}_t,\\
&=\mathbf{G}_t\mathbf{T}+\mathbf{E}_t , \quad t=(d+1),\ldots, T,
\end{split}
\end{equation*}

where
\\ $\mathbf{G}_t=\left[\begin{smallmatrix}
  F_{1}(t)Z_{t11} & \cdots &  F_{1}(t)Z_{t1(pd)} & \cdots & F_{K}(t)Z_{t11} & \cdots &  F_{K}(t)Z_{t1(pd)} \\
\svdots & \sddots & \svdots & \sddots & \svdots & \sddots & \svdots \\
  F_{1}(t)Z_{tm1} & \cdots &  F_{1}(t)Z_{tm(pd)} & \cdots & F_{K}(t)Z_{tm1} & \cdots &  F_{K}(t)Z_{tm(pd)}
\end{smallmatrix}\right]$\\
, and $\mathbf{T}=\left[\begin{smallmatrix}
\tau_{111} & \cdots & \tau_{1p1} \\ \svdots & \sddots & \svdots \\ \tau_{(pd)11} & \cdots & \tau_{(pd)p1} \\ \svdots & \sddots & \svdots \\ \tau_{11K} & \cdots & \tau_{1pK} \\ \svdots & \sddots & \svdots \\ \tau_{(pd)1K} & \cdots & \tau_{(pd)pK}\\
\end{smallmatrix}\right]$.
\section{Proof of Theorem 3.1} \label{section_proof_1}
Without loss of generality, we could rewrite Equation \ref{dynamic_lsem} as $X_{it}=\sum_{j\in \mathbf{Pa}_j^\mathcal{G}} {X}_{jt}B_{ijt}+ {E}_{it},\forall i=1,\cdots p.$, where $X_{it} $denotes the value for the $i^{th}$variable, $\mathbf{Pa}_j^\mathcal{G}$ denotes the parent nodes for the $i^{th}$variable,  $B_{ijt}$ denotes the respective coefficient between variable $i,j$ at time $t$ and $E_{it}$ represents the respective error term.\\

\subsection{Case 1: Independent non-Gaussian distribution} 
When $E_{it}$ are jointly independent and non-Gaussian, the identiability is a well-established result in the independent component
analysis \citep{shimizu}, following Corollary 13 of \citet{comon1994independent}.

\subsection{Case 2: Equal variance Gaussian distribution}
Assume that there exist two linear structural equation models as in equation \ref{dynamic_lsem} which induces $\mathcal{L}(\mathbf{X})$ with distinct graphs $\mathcal{G}$ and $\mathcal{G_0}$. 

Due to the acyclicity of the graphs, we could always find a node that has no descendants. By keeping on eliminating the nodes which have no child nodes in both $\mathcal{G}$ and $\mathcal{G_0}$ and have the same parent nodes in both $\mathcal{G}$ and $\mathcal{G_0}$, we are left with two scenarios:
\begin{itemize}
    \item There are no nodes left for both $\mathcal{G}$ and $\mathcal{G_0}$, which indicates $\mathcal{G}$=$\mathcal{G_0}$. There is a contradiction and ends the proof.
    \item There are remaining nodes, where we denote the remaining variables as $X'$, the remaining graphs as $\mathcal{G'}$ and $\mathcal{G}_0'$, the node that has no child node in $\mathcal{G'}$ as $L$, where $L$ satisfies that $PA_L^{\mathcal{G'}}\ne PA_L^{\mathcal{G}_0'}$ or $CH_L^{\mathcal{G}_0'}\ne \emptyset$, where $PA_L^{\mathcal{G'}}$ denotes the parent nodes of $L$ in graph $\mathcal{G}_0'
    $.
    \end{itemize}
    Without loss of generality, we could partition the parents of $L$ in graph $\mathcal{G}'$ into $\mathbf{Y},\mathbf{Z} ,\mathbf{W}$. Let $\mathbf{D}$ denotes the parents of $L$ in graph $\mathcal{G}_0'$ that are not adjacent to $L$ in $\mathcal{G'}$ and $\mathbf{E}$ denotes the children of $L$ in graph $\mathcal{G}_0'$ that are not adjacent to $L$ in $\mathcal{G}'$.
    
    Then we could write 
    \begin{align*}
        PA_L^{\mathcal{G}'}&=\mathbf{Y}\cup\mathbf{Z} \cup\mathbf{W},\\
        PA_L^{\mathcal{G}_0'}&=\mathbf{Z} \cup\mathbf{D}\\
        CH_L^{\mathcal{G}'}&=\emptyset\\
        CH_L^{\mathcal{G}_0'}&=\mathbf{Y}\cup\mathbf{E}
    \end{align*}
    Let $\mathbf{T}=\mathbf{Y}\cup\mathbf{W}$, consider the following two cases.
    
\textbf{1.} $\mathbf{T}=\emptyset$. Then there must exist a node $D\in\mathbf{D}$ or $E\in\mathbf{E}$, otherwise the two graphs are equivalent. If there exist a $D\in\mathbf{D}$,  based on Markov property, we have $L\indep D|(\mathbf{Z}\cup\mathbf{D}\backslash\{D\})$, which contradicts the Lemma 8 of \citet{iden_euqal_var}. If there exist a $E\in\mathbf{E}$,  based on Markov property, we have $L\indep E|(\mathbf{Z}\cup\mathbf{PA}_E^{\mathcal{G}_{0}^{'}}\backslash\{L\})$, which also contradicts the Lemma 8 of \citet{iden_euqal_var}. 

\textbf{2.}$\mathbf{T}\ne\emptyset$.
    Under the Markov condition and the causal faithfulness assumption, there exist a node $Y\in\mathbf{Y}$ that have the same set of parent nodes $\mathbf{S}$ with $L$ in both graphs, and the edge between $L$ and $Y$ reverses in the two graphs, according to Theorem 2  in \citet{chickering2013transformational}.
    Let $L^*=L|_{\mathbf{S}=s},Y*=Y|_{\mathbf{S}=s}$,for $s\in\mathbb{R}^2$. Based on Lemma 2 in \citet{peters2012identifiability}, we could have 
    \begin{align*}
        L^*=c+\beta_L Y^*+E_L,
        \end{align*} 
    where $c$ is a constant, $\beta_L,Y_L$ are the respective coefficient and the error term for variable $L$. Since $L$ has no descendants, we have $E_L\indep Y^*$, which leads to \begin{align*}
        var(L^*)=\beta_L^2var(Y^*)+\sigma^2>\sigma^2.
    \end{align*}
    Since $\mathbf{S}\supseteq \mathbf{PA}_{L}^{\mathcal{G}_{0}^{'}}$ and $det(cov(\mathbf{X}))>0$,where $det$ denotes the determinant, based on Lemma 5 in \citet{iden_euqal_var}, we could have 
    \begin{align*}  
    var(L^*)\leq \sigma^2,
    \end{align*}
    which leads to a contradiction.

\section{Proof of Theorem 5.1} \label{section_proof}

Based on the Equation \eqref{equation_amt}, we could then write
\begin{align} \label{dslem}
   \begin{cases}
    A_{t}&=\epsilon_{A_{t}}\\
    M_{t}&=\boldsymbol{\alpha}_t^T A_{t}+\mathbf{C}_t^T M_{t}+\epsilon_{M_{t}}^T\\
    &\quad + \sum_{i=1}^{d}\{\boldsymbol{\alpha}_{t-i}^T A_{t-i}+\mathbf{C}_{t-i}^T M_{t-i}+\mathbf{d}_{t-i}^TY_{t-i}\}\\
        &=(\mathbf{I}-\mathbf{C}_t^T)^{-1}[\boldsymbol{\alpha}_t^T A_{t}+\epsilon_{M_{t}}^T\\
        &\quad +\sum_{i=1}^{d}\{\boldsymbol{\alpha}_{t-i}^T A_{t-i}+\mathbf{C}_{t-i}^T M_{t-i}+\mathbf{d}_{t-i}^TY_{t-i}\}]\\
    Y_{t}&=\gamma_t A_{t}+ \boldsymbol{\beta}_t^T M_{t}+\epsilon_{Y_{t}}\\
    &\quad +\sum_{i=1}^{d}\{\gamma_{t-i} A_{t-i}+ \boldsymbol{\beta}_{t-i}^T M_{t-i}+f_{t-i}Y_{t-i}\}
\end{cases}
\end{align}

As mentioned in Section \ref{section_causal_effect}, we could decompose the data matrix $X_t$ into the treatment variable, $A_t$, mediator variable $M_t$ and the outcome variable $Y_t$, $X_t=[A_t,M_t^T,Y_t]$.
 
Let overbar denotes the history of the respective variables, e.g. $\overline{A}_t=(A_1,\cdots, A_t)$, $Y_t{(\overline{a}_{t-1})}$ denote the potential outcome at time $t$ if the individual had the treatment history $\overline{a}_{t-1}$, and $H_t$ denotes the history up to time $t$, $H_t=(\overline{M}_{t},\overline{Y}_{t},\overline{A}_{t})$.

 Under Assumptions in Section \ref{assumption} and Equation \eqref{dslem}, the dynamic causal effect could be written as

\begin{align*}
&\mathbb{E}[Y_{t+1}^{(\overline{a}_{t},a)}|H_t])-\mathbb{E}[Y_{t+1}^{(\overline{a}_{t},0)}|H_t])
    \\
    &=\mathbb{E}[Y_{t+1}^{(\overline{a}_{t},a)}|\overline{M}_{t}=\overline{m}_{t},\overline{Y}_{t}=\overline{y}_{t},\overline{A}_{t}=\overline{a}_{t}])
    \\
    &\qquad -\mathbb{E}[Y_{t+1}^{(\overline{a}_{t},0)}|\overline{M}_{t}=\overline{m}_{t},\overline{Y}_{t}=\overline{y}_{t},\overline{A}_{t}=\overline{a}_{t}])
    \\
    &=\mathbb{E}[Y_{t+1}^{(\overline{a}_{t},a)}|\overline{M}_{t}=\overline{m}_{t},\overline{Y}_{t}=\overline{y}_{t},\overline{A}_{t}=\overline{a}_{t}, A_{t+1}=a])
    \\
    &\qquad -\mathbb{E}[Y_{t+1}^{(\overline{a}_{t},0)}|\overline{M}_{t}=\overline{m}_{t},\overline{Y}_{t}=\overline{y}_{t},\overline{A}_{t}=\overline{a}_{t}, A_{t+1}=0])
    \\
    &=\mathbb{E}[Y_{t+1}|\overline{M}_{t}=\overline{m}_{t},\overline{Y}_{t}=\overline{y}_{t},\overline{A}_{t}=\overline{a}_{t}, A_{t+1}=a])
    \\
    &\qquad -\mathbb{E}[Y_{t+1}|\overline{M}_{t}=\overline{m}_{t},\overline{Y}_{t}=\overline{y}_{t},\overline{A}_{t}=\overline{a}_{t}, A_{t+1}=0])\\
    &=\mathbb{E} [\gamma_{t+1} A_{t+1}+ \boldsymbol{\beta}_{t+1}^T M_{t+1}+\epsilon_{Y_{t+1}}\\
    &\quad +\sum_{i=1}^{d}\{\gamma_{t+1-i} A_{t+1-i}+ \boldsymbol{\beta}_{t+1-i}^T M_{t+1-i}+f_{t+1-i}Y_{t+1-i}\}\\
    & \qquad   \qquad |\overline{M}_{t}=\overline{m}_{t},\overline{Y}_{t}=\overline{y}_{t},\overline{A}_{t}=\overline{a}_{t}, A_{t+1}=a]\\
    &\quad -\mathbb{E} [\gamma_{t+1} A_{t+1}+ \boldsymbol{\beta}_{t+1}^T M_{t+1}+\epsilon_{Y_{t+1}}\\
    &\quad +\sum_{i=1}^{d}\{\gamma_{t+1-i} A_{t+1-i}+ \boldsymbol{\beta}_{t+1-i}^T M_{t+1-i}+f_{t+1-i}Y_{t+1-i}\}\\
    &\qquad \qquad   \qquad |\overline{M}_{t}=\overline{m}_{t},\overline{Y}_{t}=\overline{y}_{t},\overline{A}_{t}=\overline{a}_{t}, A_{t+1}=0]
        \end{align*}
    \begin{align*}
    &=\mathbb{E}[\gamma_{t+1} A_{t+1}+ \beta_{t+1}^T M_{t+1}|\overline{M}_{t}=\overline{m}_{t},\overline{Y}_{t}=\overline{y}_{t},\overline{A}_{t}=\overline{a}_{t}, A_{t+1}=a]\\
    &\quad -\mathbb{E}[\gamma_{t+1} A_{t+1}+ \beta_{t+1}^T M_{t+1}|\overline{M}_{t}=\overline{m}_{t},\overline{Y}_{t}=\overline{y}_{t},\overline{A}_{t}=\overline{a}_{t}, A_{t+1}=0]\\
    &=\mathbb{E}[(\gamma_{t+1}+\beta_{t+1}^T(\mathbf{I}-\mathbf{C}_{t+1}^T)^{-1}\boldsymbol{\alpha}_{t+1}^T) A_{{t+1}}|H_t, A_{t+1}=a)]\\
    &\qquad -\mathbb{E}[(\gamma_{t+1}+\beta_{t+1}^T(\mathbf{I}-\mathbf{C}_{t+1}^T)^{-1}\boldsymbol{\alpha}_{t+1}^T) A_{{t+1}}|H_t, A_{t+1}=0)]\\
    &=(\gamma_{t+1}+\beta_{t+1}^T(\mathbf{I}-\mathbf{C}_{t+1}^T)^{-1}\boldsymbol{\alpha}_{t+1}^T)a,
\end{align*}
where the second equality is due to the sequential ignorability assumption, the third equality is due to the consistency assumption, the fourth and fifth equality are due to the Equation \eqref{dslem}.

\section{Details on the spline selection}
\label{spline_detail}
A B-spline with the order of zero is defined as follows \citep{basis}:
\begin{align*}
    f_{i,0}(x)&=
    \begin{cases}
     0 \qquad \text{if } x< k_{(i-1)} \text{or } x\geq k_{i},\\
     1 \qquad \text{otherwise},
    \end{cases}\\
    & i=1,\cdots,k,
\end{align*}
where $f_{i,0}(x)$ denotes $i^{th}$ B-spline function of order zero, $k_i$ denotes the $i^{th}$ knots that characterize the B-spline, $k$ denotes the number of the knots.

\begin{figure}[h] 
\centering
\includegraphics[width=0.8\linewidth]{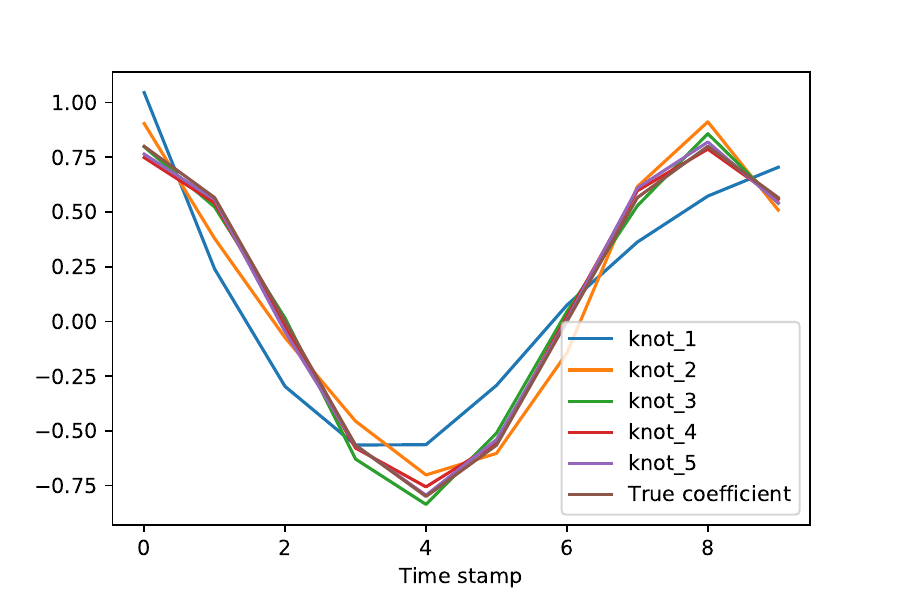}
\caption{Causal strength estimates with different number of knots under scenario 1 with cosine function in the dynamic LSEM setup using the proposed method.}
\label{basis}
\end{figure}

Then, B-splines with higher orders could be recursively defined as follows:
\begin{align*}
    f_{i,r}(x)=v_{i,r}f_{i,r-1}(x)+(1-v_{i+1,r})f_{i+1,r-1}(x),
\end{align*}
where $r$ denotes the degree of the basis function, and 
\begin{align*}
    v_{i,r}=
    \begin{cases}
     \frac{x-k_{i-1}}{k_{i+r-1}-k_{i-1},} \qquad k_{i+r-1}\ne k_{i-1},\\
     0 \qquad \text{otherwise}.
    \end{cases}
\end{align*}
As shown in Figure \ref{basis}, increasing the number of knots from 1 to 2 would lead to better estimates for the causal strength, while similar estimates would be obtained when there are more than 2 knots.

\section{Details on Variational Autoencoder (VAE)} \label{vae}
As discussed in Section \ref{identification}, we have assumed that the latent variables follow a normal distribution and assumed the underlying the data generation mechanism is the dynamic LSEM/SVAR model as described in Equation \ref{dynamic_lsem},\ref{equation_SVAR}, when applying the variatioal autoencoder. 
 As shown in Section \ref{section_basis}, after the basis transformation, Equation \ref{lsem_final},\ref{equation_final} could be reformulated into a general form \begin{align*}
   \Tilde{\textbf{X}}=\Tilde{\textbf{E}}\Tilde{\textbf{
A}} ,i.e., \Tilde{\textbf{X}}^T=\Tilde{\textbf{
A}}^T\Tilde{\textbf{E}}^T,
 \end{align*} where $\Tilde{\textbf{X}}$
 contains the known data matrix, $\Tilde{\textbf{E}}$
 is the error matrix and the $\Tilde{\textbf{A}}$ 
 contains the coefficients that we want to measure, such as $\boldsymbol{\Gamma}$
. 

Then, we have the following model:
\begin{align*}
    \text{Encoder:}\quad &\Tilde{\textbf{X}}^T=f_2(\Tilde{\textbf{A}}^T)f_1(\Tilde{\textbf{E}}^T)\\
    \text{Decoder:}\quad &\Tilde{\textbf{E}}^T=f_3(\Tilde{\textbf{A}}^{-1T})f_4(\Tilde{\textbf{X}}^T),\\
\end{align*}
where $f_1,f_4$
 are the identity mapping and $f_2,f_3$
 is the MLP layer \citet{DAGGNN}. In this way, we could use the VAE to estimate the model parameters by minimizing the loss function to get the estimate for $\Tilde{\textbf{A}}$
. The MLP layer here adds non-linearly to the model so it could be generalized to nonlinear SEM. We utilize normal distribution for the encoder and decoder of VAE.

\section{Additional real data analysis}
\begin{figure}[h]
    \centering
    \includegraphics[width=0.8\linewidth]{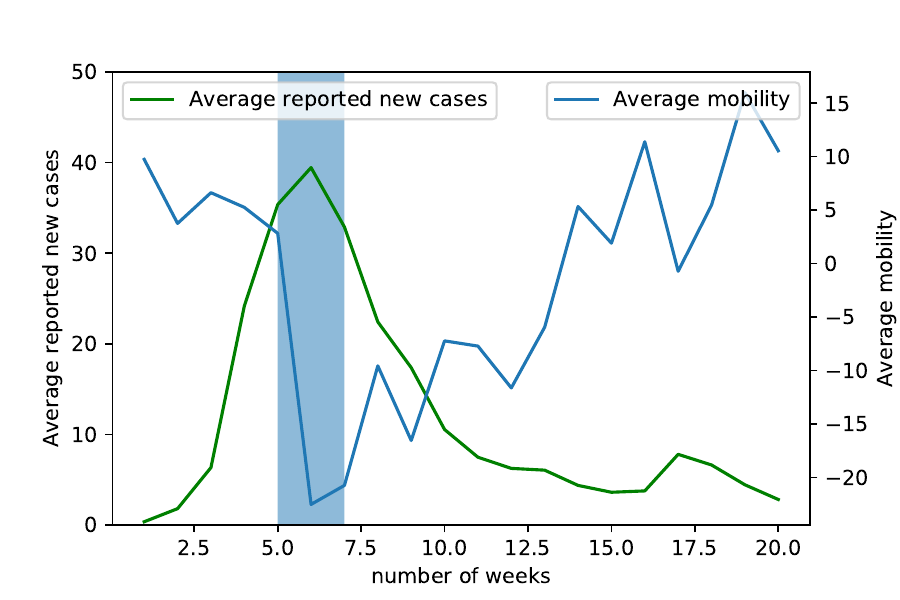}
    \caption{Average reported new cases and mobility with respect to time. The shaded region represents the time period that the contact restriction policy started to implement.}
    \label{real_data}
\end{figure}

Figure \ref{real_data} plots how the average reported new COVID-19 cases and the average mobility index changes with time, where the shaded region representes the time period that the contact restriction policy begin to implement. We could see from the figure that, after the contact restriction policy beginning to implement, people tend to travel less since the mobility variable has a sudden drop. Also, the average reported new cases shows a decreasing trend after the policy begins to implement.

\section{Additional results on simulation studies}
\label{addtional}
In this section, we provide some additional numerical results for the simulation studies.\\

Table \ref{table_lag_W_t} presents the metrics for the estimated time-lagged weight matrix in the dynamic SVAR model setup. Since both ANOCE and NOTEARS are not capable to capture the time-lagged dependency, we compare the performance of the proposed method with DYNOTEARS. Also, since $\mathbf{W}_t$ is not time-varying in the simulation, we omit the MSE metric in the table. As shown in Table \ref{table_lag_W_t}, our method outperforms the DYNOTEARS for all three metrics in both scenarios.

\begin{table}[H]

\caption{The empirical comparison of estimated time-lagged weight matrix $\mathbf{W}_t$ in the dynamic SVAR model setup. The number in parenthesis denotes the standard deviation. }
\centering
\begin{tabular}{cccc@{}c@{}c@{}c@{}}
\toprule
Methods & Metric & F1          & F2                     &  &  \\
\midrule
\multirow{4}{*}{DYNOTEARS} & FDR    & 0.40 (0.01) & 0.36(0.01)     &    \\
& TPR    & 0.27(0.01)    & 0.27(0.01)    &    \\
& SHD    & 3.64(0.02)  & 3.75(0.02)       &    \\ 
\midrule
\multirow{4}{*}{Proposed}    & FDR    & \textbf{0.28(0.02)}   & \textbf{0.26(0.01)}        &    \\
& TPR    & \textbf{0.95(0.03)}    & \textbf{0.98(0.00)}     &  & \\
 & SHD    & \textbf{2.20(0.01)}   & \textbf{1.93(0.08)}       &    \\

\bottomrule
\end{tabular}
\label{table_lag_W_t}
\end{table}

\begin{figure*}[h]
\begin{subfigure}{0.9\columnwidth}
  \centering
  \includegraphics[width=\columnwidth]{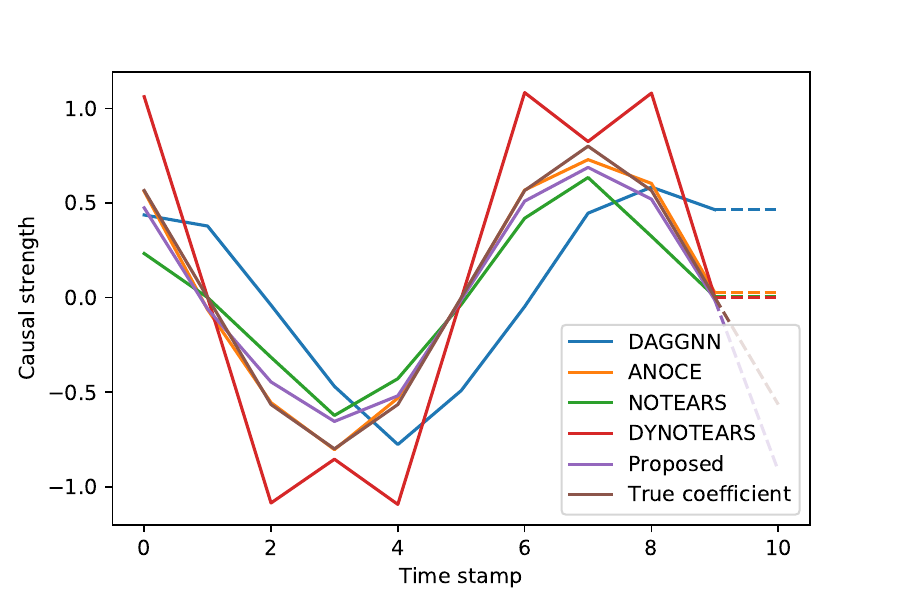}
  \caption{Cosine function F1}
\end{subfigure}%
\hfill
\begin{subfigure}{0.9\columnwidth}
  \centering
  \includegraphics[width=\columnwidth]{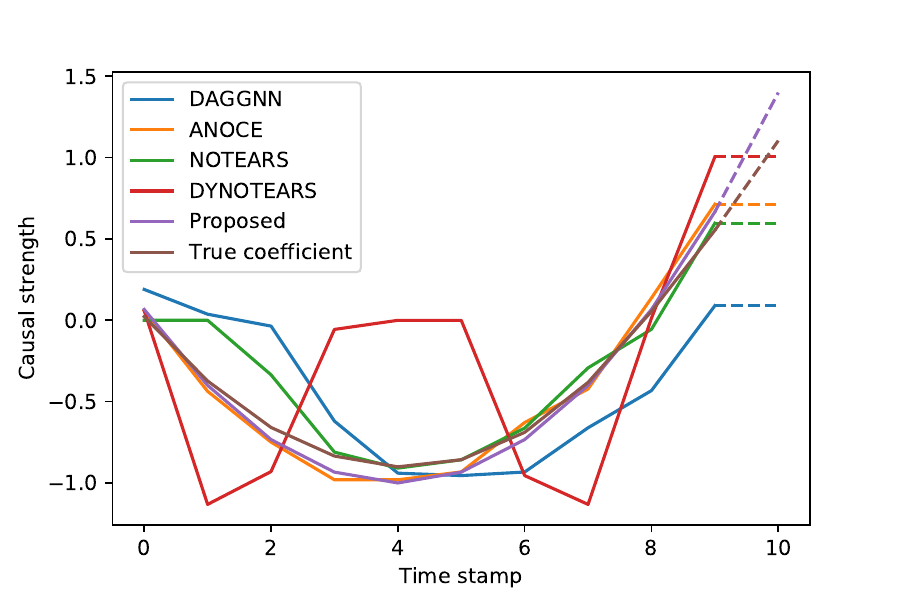}
  \caption{Quadratic function F2}
\end{subfigure}
\caption{Estimated causal strength using the proposed method and benchmark methods in the dynamic SVAR model setup. The solid line denotes the estimated coefficient in the ten time stamps and the dashed line denotes the one-step ahead prediction.}
\label{figure_lag}
\end{figure*}

Figure \ref{figure_lag} plots the estimated casual strength in the dynamic SVAR setup. We could see that, the proposed method could provide more accurate estimates on the past data and also is able to make predictions that more close to the true value.
 
\begin{figure*}[h]
    \begin{subfigure}{\columnwidth}
    \includegraphics[width=\columnwidth]{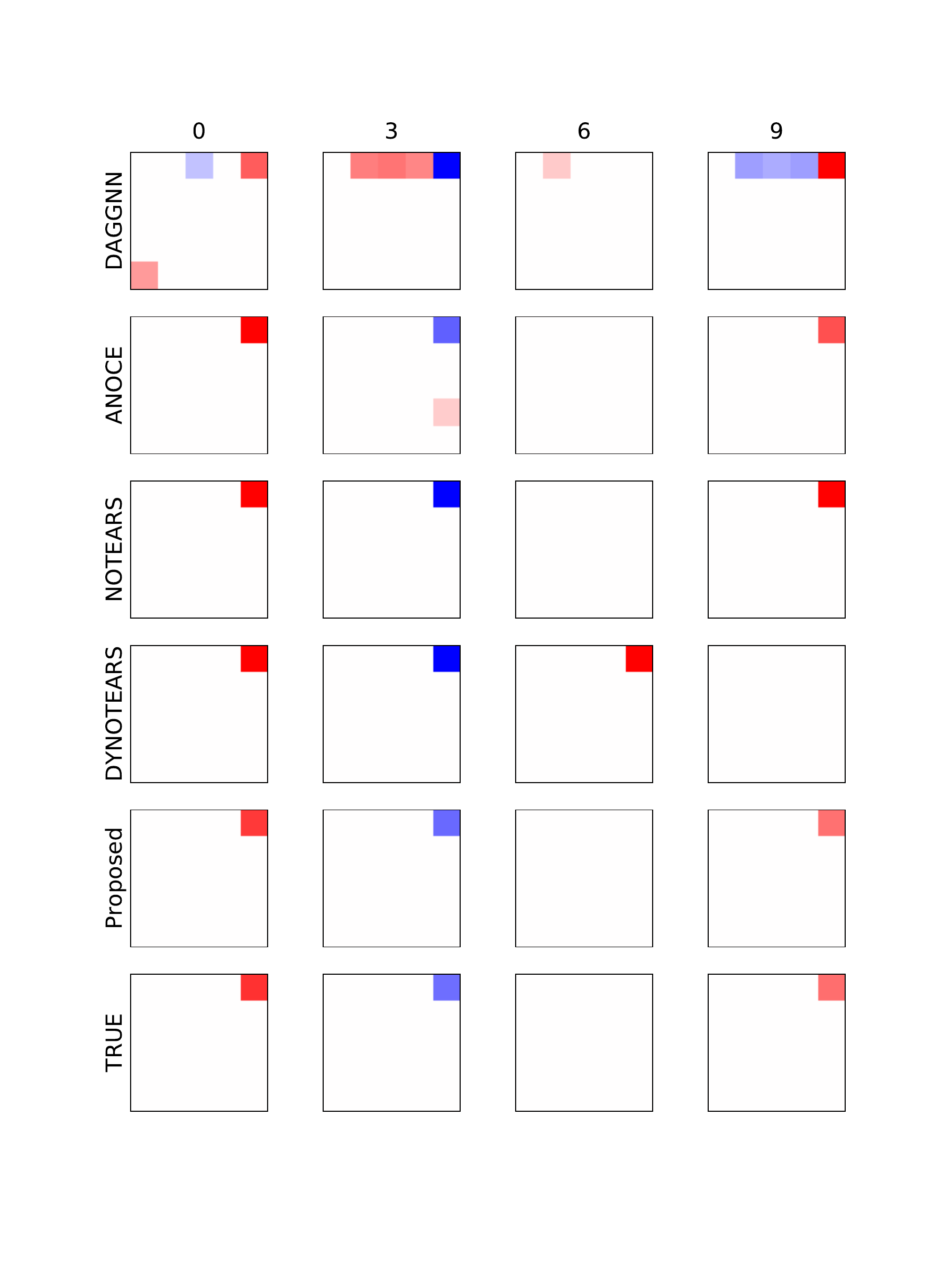}
  \caption{Cosine function F1}
\end{subfigure}%
  \hfill
    \begin{subfigure}{\columnwidth} \includegraphics[width=\columnwidth]{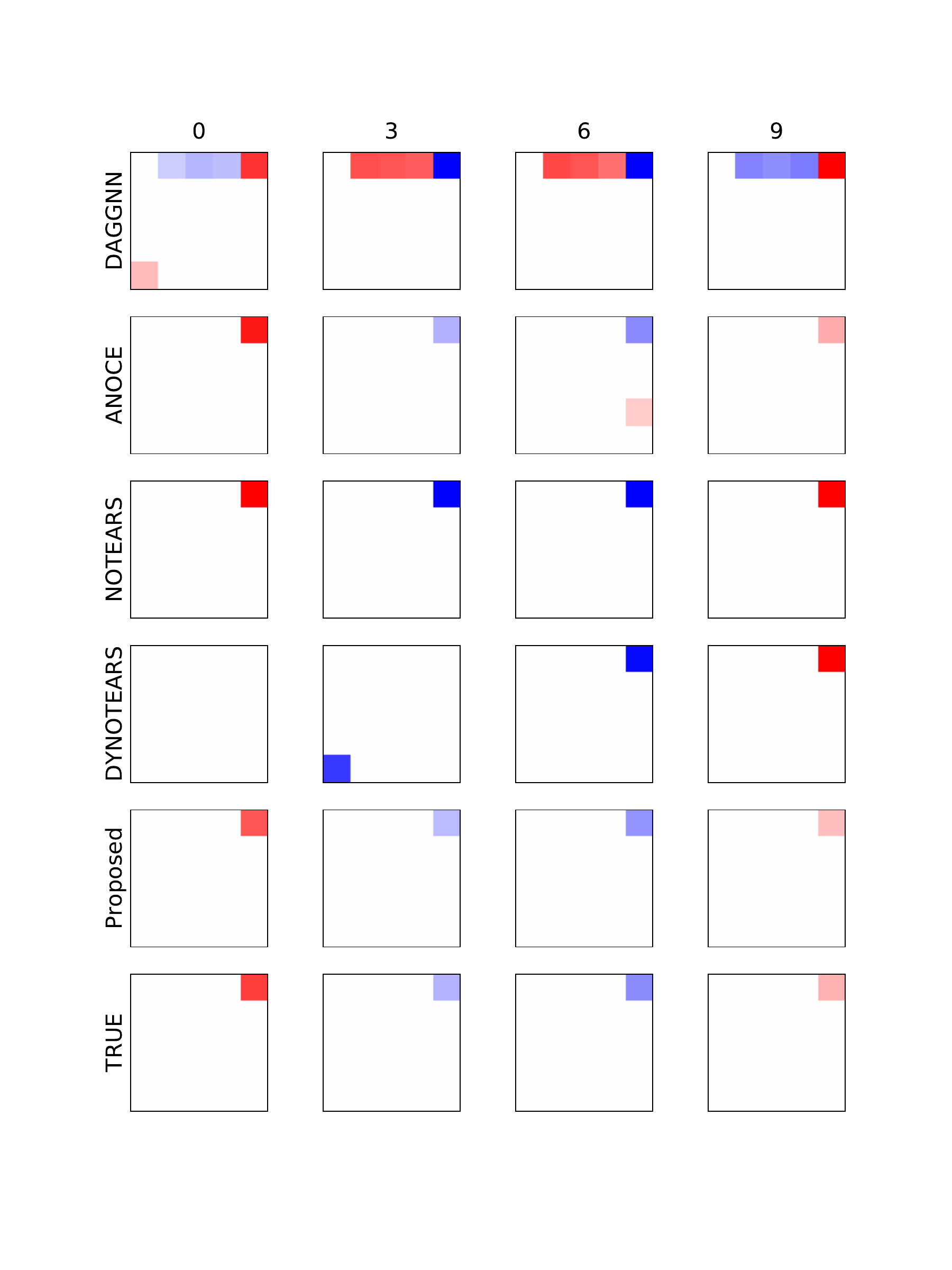}
  \caption{Quadratic function F2}
\end{subfigure}%
    \caption{Estimated causal graphs at multiple time-stamps using the proposed
method and the benchmarks for Scenario 1 based on the dynamic LSEM setup.}
\label{S1_nolag_graph}
\end{figure*}

\begin{figure*}[h]
    \begin{subfigure}{\columnwidth}
\includegraphics[scale=0.3]{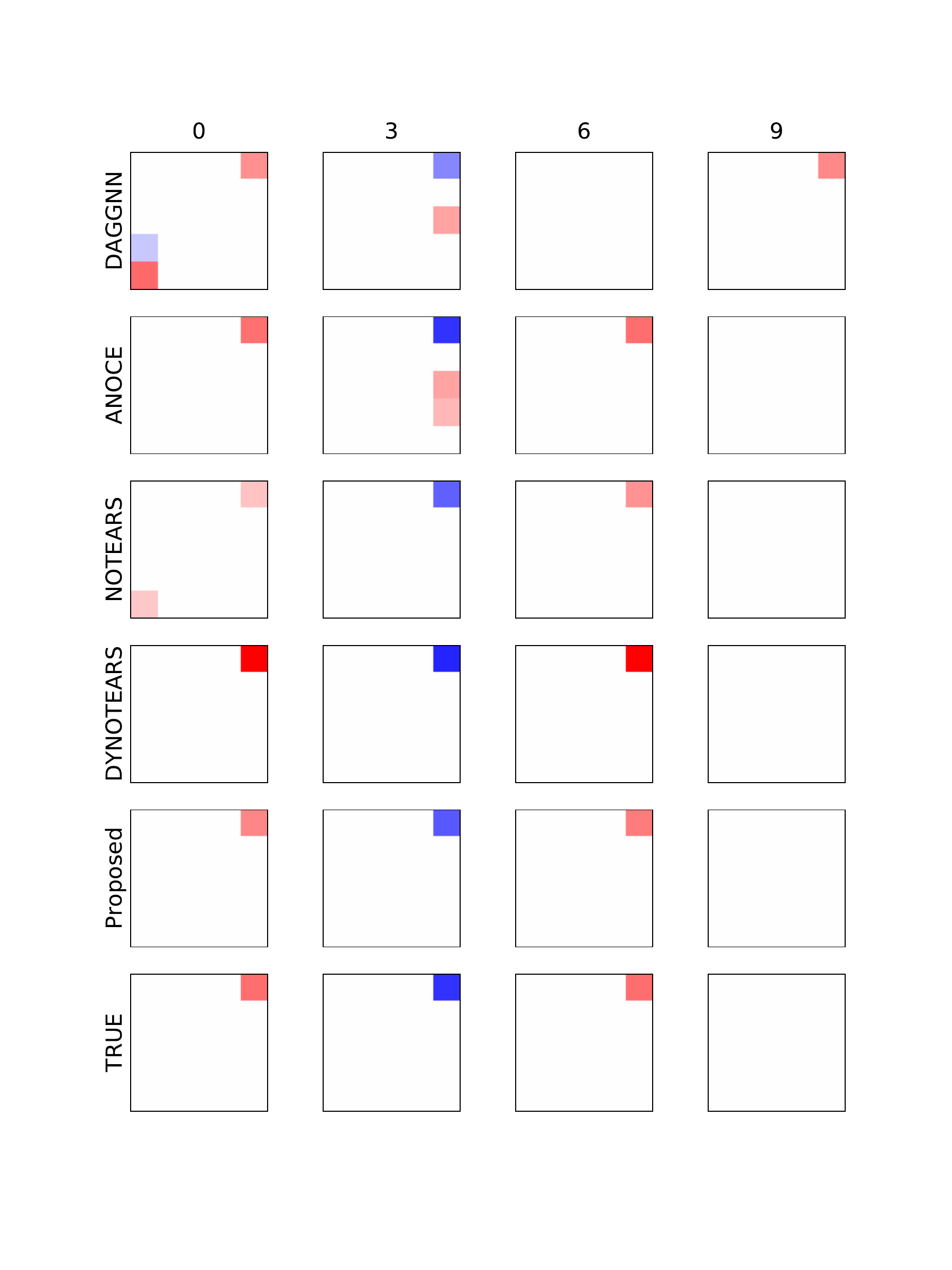}
  \caption{Cosine function F1}
\end{subfigure}%
    \begin{subfigure}{\columnwidth} \includegraphics[scale=0.3]{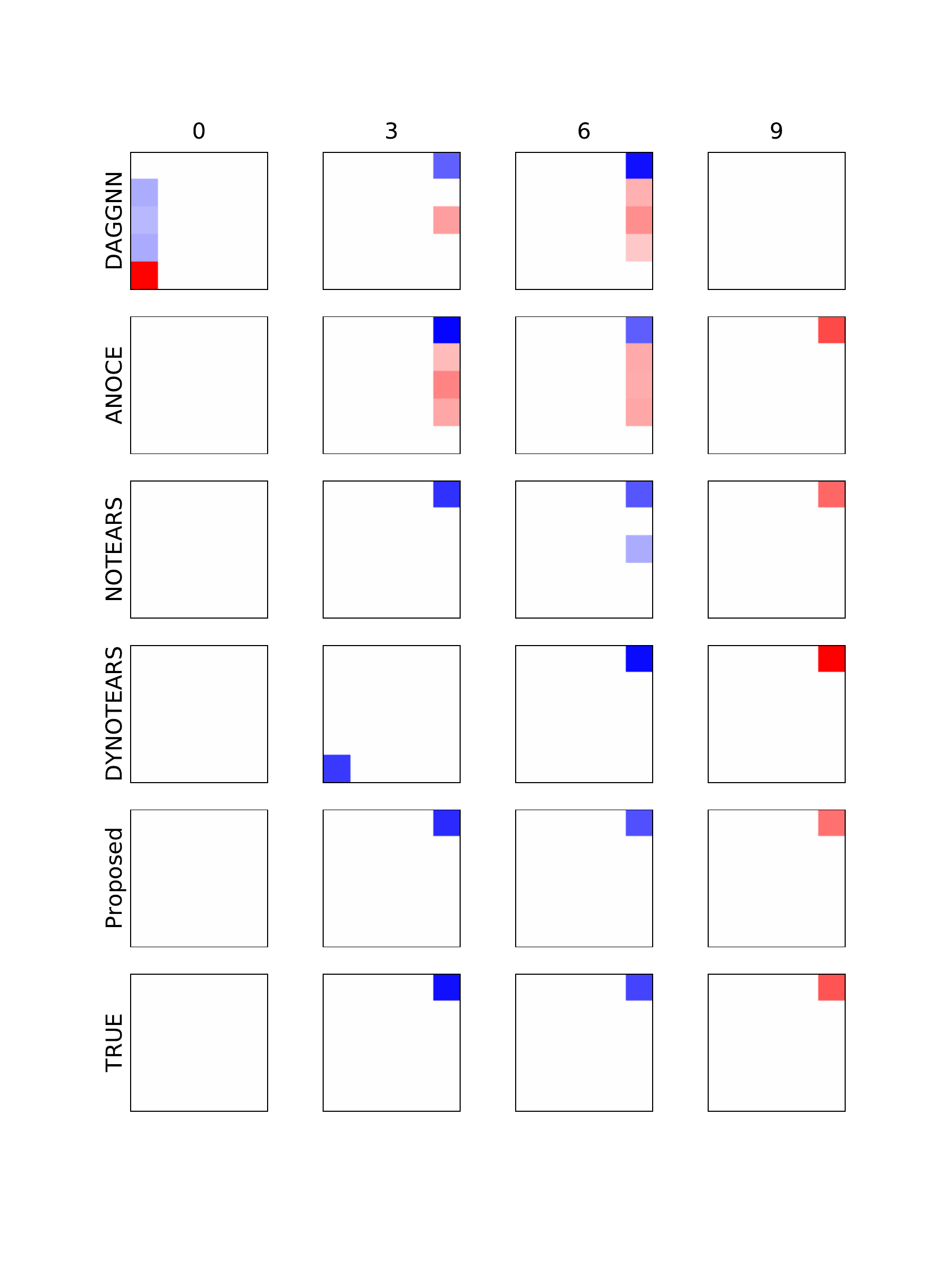}
  \caption{Quadratic function F2}
\end{subfigure}%
    \caption{Estimated causal graphs at multiple time-stamps using the proposed
method and the benchmark based on the dynamic SVAR model setup.}
\label{lag_graph}
\end{figure*}

 Moreover, we provide the estimated graphs of each method for the simulated data at multiple time-stamps.
 Figure \ref{S1_nolag_graph} plot the estimated graph for each method for the dynamic LSEM setup. Figure \ref{lag_graph} plots the estimated graph for each method for the dynamic SVAR setup. As shown in these figures, our method could estimate the underlying dynamic causal graph better.
 
\begin{table}[h]
\caption{The empirical comparison of the estimated causal graph for the synthetic data in the dynamic LSEM setup, when $m=1, T=100$. The number in parenthesis denotes the standard deviation. }
\begin{tabular}{llllll}
\toprule
Metric & CD-NOD          & Proposed                 &  &  \\
\midrule
 FDR    & 0.77(0.02)   & \textbf{0.01 (0.00)}      &  &  \\
 TPR    & 0.34(0.03)     & \textbf{0.94(0.00)}    &  &  \\
  SHD    & 3.09(0.22)    & \textbf{0.07(0.00)}      &  &  \\
 MSE    & 0.41(0.01),           & \textbf{0.03(0.00)}         &  &  \\

\bottomrule
\end{tabular}
\label{table_m=1}
\end{table}

 Table \ref{table_m=1} shows the empirical results of the estimated causal graph for the synthetic data in the dynamic LSEM setup, when $m = 1,T = 100.$ The causal strength changes with time following $f(t)=cos(\frac{t}{30\pi})*0.8$. Since the other methods are not designed for no-replicate scenarios, we only present the results for the proposed method and CD-NOD \citep{zhang}. As shown in this table, our method has shown superior performance in all metrics, indicating its ability to correctly estimate the dynamic causal structure with long-time and no-replicate scenario.

 Table \ref{table_vary} shows the empirical results of the estimated causal graph for the synthetic data in the dynamic LSEM setup, for scenario S1F1, when the graph sizes vary. We could see from the table that as the number of nodes increases, the proposed method still have superior performance among all the metrics. It shows the proposed methods' generality to larger scale of graphs.
 
\begin{table}[h]
\caption{The empirical comparison of the estimated causal graph for the synthetic data in the dynamic LSEM setup for scenario S1F1,when the number of nodes varies. The number in parenthesis denotes the standard deviation. }
\centering
\begin{tabular}{c@{}cc@{}cc@{}cc@{}}
\toprule
Methods & Metric & p=5   & p=8         & p=10                    &  &  \\
\midrule
\multirow{4}{*}{DAGGNN}    & FDR    & 0.83(0.01) &0.31(0.08)  & 0.80(0.01)       &  &  \\
& TPR    & 0.82(0.04)  &0.34(0.04)  & 0.90(0.03)     &  &  \\
 & SHD    & 3.36(0.16)  &3.32(0.70)  & 3.38(0.15)        &  &  \\
& MSE    & 0.65(0.04) & 0.48(0.04)       & 2.69(0.26)               &  &  \\
\midrule
\multirow{4}{*}{ANOCE}    & FDR    & 0.54(0.03)  & 0.62(0.02)  & 0.60(0.04)       &  &  \\
& TPR    & \textbf{1.00(0.00)}    & \textbf{1.00(0.00)} & \textbf{1.00(0.00)}    &  &  \\
 & SHD    & 1.83(0.13) & 3.07(0.24)   & 4.23(0.41)        &  &  \\
& MSE    & 0.06(0.02),  &  \textbf{0.04 (0.01)}      & \textbf{0.00 (0.00)}                &  &  \\
\midrule
\multirow{4}{*}{NOTEARS}    & FDR    & \textbf{0.00(0.00)}   & \textbf{0.00(0.00)}    & \textbf{0.00(0.00)}   &  &  \\
& TPR    & \textbf{1.00(0.00)}    & \textbf{1.00(0.00)} & \textbf{1.00(0.00)})    &  &  \\
 & SHD    & \textbf{0.00(0.00)}    & \textbf{0.00(0.00)}    & \textbf{0.00(0.00)}   &  &  \\
& MSE    & 0.37(0.00),        & 0.37 (0.00)        & 0.37(0.00)        &  &  \\
\midrule
\multirow{4}{*}{DYNOTEARS}    & FDR    & \textbf{0.00(0.00)} & \textbf{0.00(0.00)}  & 0.70(0.00)        &  &  \\
& TPR    & \textbf{1.00(0.00)}  & \textbf{1.00(0.00)} & 0.30(0.00)     &  &  \\
 & SHD    & \textbf{0.00(0.00)}  & \textbf{0.00(0.00)}  & 1.40(0.00)       &  &  \\
& MSE    & 0.66(0.00),        & 0.66(0.00)        & 0.66(0.01)        &  &  \\
\midrule
\multirow{4}{*}{Proposed} & FDR    & \textbf{0.00 (0.00)} & \textbf{0.00 (0.00)}& \textbf{0.01   (0.00)}     &  &  \\
& TPR    & \textbf{1.00(0.00)}    & \textbf{0.99   (0.01)}& \textbf{0.99   (0.01)}    &  &  \\
& SHD    & \textbf{0.01(0.01)}& \textbf{0.02(0.01)}& \textbf{0.04(0.02)}         &  &  \\
& MSE    & \textbf{0.00(0.00)}  & \textbf{0.04(0.01)}& {0.10 (0.03)}                        &  & \\
\bottomrule
\end{tabular}
\label{table_vary}
\end{table}

\end{document}